%% file: main.tex
\documentclass[letterpaper, 10 pt, conference]{ieeeconf}

\IEEEoverridecommandlockouts                              

\usepackage{graphics} 
\usepackage{epsfig} 
\usepackage{mathptmx} 
\usepackage{times} 
\usepackage{amsmath} 
\usepackage{amssymb}  
\usepackage{xspace}
\usepackage[caption=false]{subfig}
\usepackage{threeparttable}

\usepackage{color} 

\title{\LARGE \bf
HR-APR: APR-agnostic Framework with Uncertainty Estimation and Hierarchical Refinement for Camera Relocalisation}

\author{Changkun Liu$^{1}$, Shuai Chen$^{3}$, Yukun Zhao$^{1}$, Huajian Huang$^{1}$, Victor Prisacariu$^{3}$ and Tristan Braud$^{1,2}$
\thanks{$^{1}$ Authors are with the Department of Computer
Science and Engineering, The Hong Kong University of Science and Technology, Hong Kong,
        {\tt\small \{cliudg,yzhaoeg,hhuangbg\}@connect.ust.hk}, {\tt\small braudt@ust.hk}}%
\thanks{$^{2}$ Tristan Braud is also with the Division of Integrated Systems Design, The Hong Kong University of Science and Technology, Hong Kong
}%
\thanks{$^{3}$ Shuai Chen and Victor Prisacariu are with the Active Vision Laboratory, Department of Engineering Science, University of Oxford, United Kingdom, {\tt\small \{shuaic,victor\}@robots.ox.ac.uk}}
}%

\newcommand{\sysname}{HR-APR\xspace}

\begin{document}

\maketitle
\thispagestyle{empty}
\pagestyle{empty}

\begin{abstract}
Absolute Pose Regressors (APRs) directly estimate camera poses from monocular images, but their accuracy is unstable for different queries. Uncertainty-aware APRs provide uncertainty information on the estimated pose, alleviating the impact of these unreliable predictions. However, existing uncertainty modelling techniques are often coupled with a specific APR architecture, resulting in suboptimal performance compared to state-of-the-art (SOTA) APR methods.  This work introduces a novel APR-agnostic framework, \sysname, that formulates uncertainty estimation as cosine similarity estimation between the query and database features. It does not rely on or affect APR network architecture, which is flexible and computationally efficient. In addition, we take advantage of the uncertainty for pose refinement to enhance the performance of APR. The extensive experiments demonstrate the effectiveness of our framework, reducing 27.4\% and 15.2\% of computational overhead on the 7Scenes and Cambridge Landmarks datasets while maintaining the SOTA accuracy in single-image APRs.

\end{abstract}

\input{Sections/1Introduction}

\input{Sections/2Relatedwork}

\input{Sections/3Method}

\input{Sections/4Experiment}

\input{Sections/5Conclusion}









\bibliographystyle{IEEEtran} 
\bibliography{IEEEabrv,IEEEexample}
\end{document}

%% file: Sections/1Introduction.tex
\section{INTRODUCTION}
Camera relocalisation, estimating the six degrees of freedom (6-DoF) absolute camera pose in the world space, is a core component in many applications, including mobile robotics, navigation, and augmented reality. 
In recent years, Absolute Pose Regressors (APRs) have emerged as an appealing approach for monocular camera pose estimation. APRs use neural networks for directly inferring 6DoF camera poses from monocular frames and offer advantages in terms of  computation and memory footprint over classical 3D structure-based methods~\cite{dusmanu2019d2,sarlin2019coarse,taira2018inloc,noh2017large,sattler2016efficient}. However, APR methods often struggle with generalization beyond their training data, leading to inaccurate predictions~\cite{sattler2019understanding}.

\begin{figure}[t]
  \centering
  \includegraphics[width=.4\textwidth]{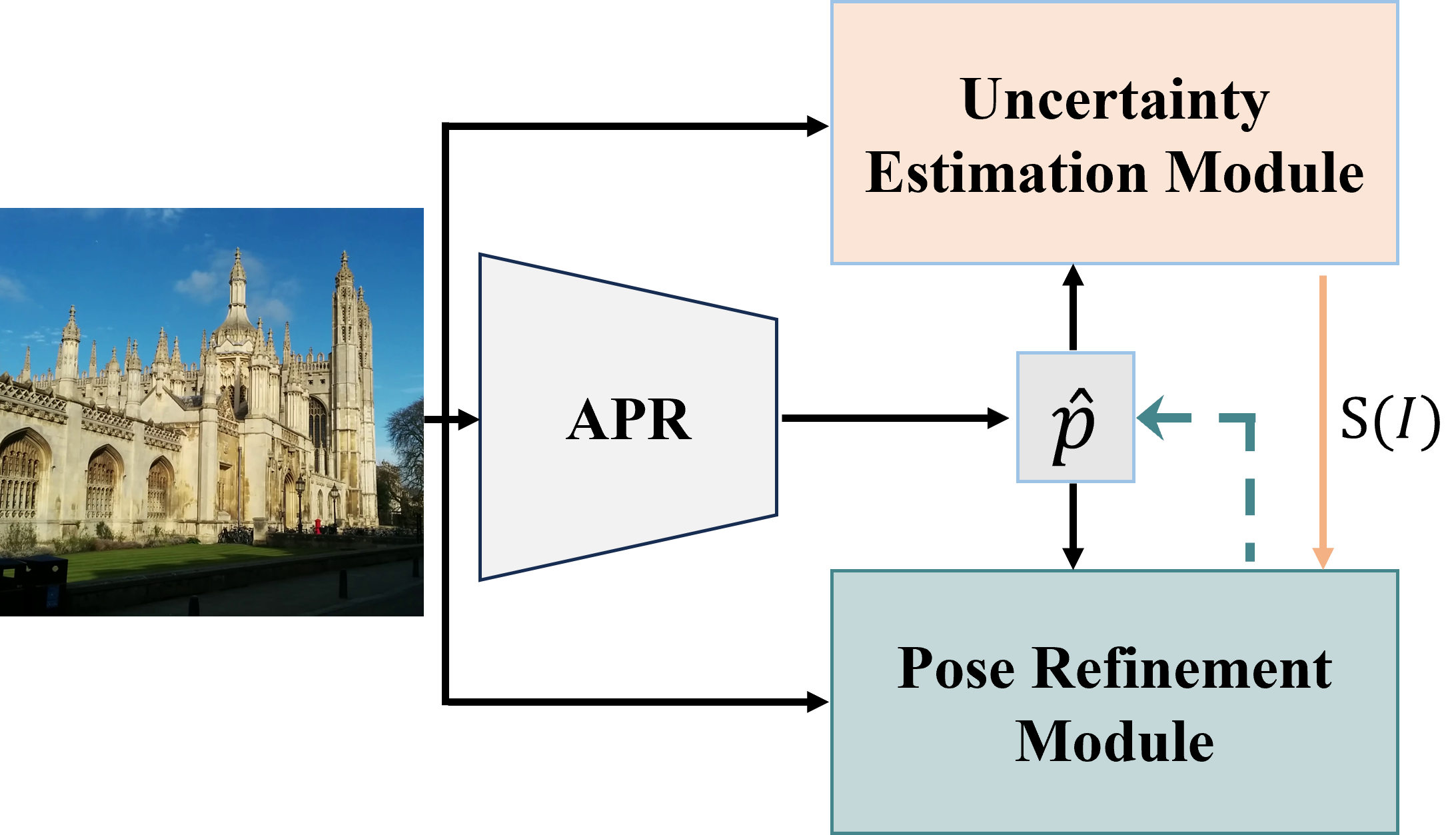}
  \caption{HR-APR: APR-agnostic Framework with Uncertainty Estimation and Hierarchical Refinement.}
\label{fig:framework}
\end{figure}


Following the first APR model, PoseNet~\cite{kendall2015posenet}, numerous methods have been proposed to enhance the robustness and accuracy of pose predictions, including modifications to network architectures~\cite{walch2017image,melekhov2017image,wu2017delving, wang2019atloc, shavit2021learning},  augmentation of the training set with labelled synthetic images~\cite{naseer2017deep,wu2017delving,sattler2019understanding}, different training strategies and loss functions~\cite{kendall2017geometric, chen2021direct,chen2022dfnet}.
Although these methods improve accuracy, they do not distinguish inaccurate predictions.
In this study, we first re-examine prevailing APRs~\cite{kendall2015posenet,shavit2021learning,chen2022dfnet} and demonstrate that even such state-of-the-art (SOTA) APR methods 
still output significantly unreliable poses, raising the need for uncertainty estimation in real-life applications. 


Uncertainty-aware (UA) APRs aim to distinguish unreliable predictions by providing additional uncertainty information with the estimated pose. 
Bayesian PoseNet~\cite{kendall2016modelling} models the uncertainty by generating multiple hypotheses for each image at inference time. In AD-PoseNet~\cite{huang2019prior}, uncertainty is quantified via prior guided dropout. CoordiNet~\cite {moreau2022coordinet} models heteroscedastic uncertainty during training.  Deng \textit{et al.}~\cite{deng2022deep,bui20206d} and Zangeneh \textit{et al.}~\cite{10160466} use pose distributions to represent uncertainty. However, these methods can be time-consuming~\cite{kendall2016modelling,huang2019prior}, and have weak extensibility as they rely on specific network architectures and specific training schemes~\cite{bui20206d,10160466,huang2019prior,moreau2022coordinet}. Furthermore, although these UA-APRs provide both pose predictions and uncertainty estimates, the accuracy of estimate poses is much lower than SOTA non-UA APRs~\cite{shavit2021learning,chen2022dfnet} that only output poses.

Such a gap between accuracy improvement and uncertainty estimation raises the need for more modular and more generic approaches to uncertainty estimation. In this paper, we propose \sysname, a novel test-time APR-agnostic framework with uncertainty estimation and pose refinement (see Figure~\ref{fig:framework}). The uncertainty estimation module integrates a new pose-based retrieval algorithm that fetches image feature embeddings in the training set. The cosine similarity between these retrieved features and the query image is then calculated to measure uncertainty. 
We leverage the uncertainty to optimize an iterative pose refinement pipeline~\cite{chen2023refinement}. 
 
 We summarize our main contributions as follows:
 



\begin{enumerate}
    \item  We propose a novel APR-agnostic uncertainty estimation module to predict the uncertainty of the APR output during test time. This module integrates a new pose retrieval algorithm that fetches image feature embeddings in the training set. The cosine similarity between these retrieved features and the query image is then calculated to measure uncertainty.

    \item  We evaluate the accuracy of our uncertainty estimation module over three APR models with different architectures on indoor and outdoor datasets. The proposed  method displays a clear correlation between pose error and uncertainty with similar performance across APR models, demonstrating the validity of the approach.
    
    \item We further leverage the predicted uncertainty to reduce the overhead of an iterative pose refinement algorithm. \sysname reduces the SOTA APR refinement pipeline's overhead by 27.4\% and 15.2\% on the indoor and outdoor datasets, respectively, while maintaining the SOTA accuracy of single-image APR methods.
    
    
\end{enumerate}

%% file: Sections/2Relatedwork.tex
\section{RELATED WORK}
\subsection{Absolute Pose Regression}
\label{subsec:apr}
APRs train neural networks for regressing the 6-DoF camera pose of query images. The seminal work in this area is introduced by PoseNet (PN)~\cite{kendall2015posenet}. Further modifications to network architecures~\cite{walch2017image,melekhov2017image,wu2017delving, wang2019atloc, shavit2021learning} and different training strategies ~\cite{kendall2016modelling,kendall2017geometric} have been made to improve accuracy and performance. MS-Transformer (MS-T)~\cite{shavit2021learning} follows MS-PN~\cite{blanton2020extending}, which extends the single-scene APRs to multiple-scene APRs. Other approaches localise from image sequences~\cite{brahmbhatt2018geometry,clark2017vidloc,radwan2018vlocnet++,valada2018deep}. However, ~\cite{sattler2019understanding}  has shown that most APR methods 
do not generalize well beyond training set via image retrieval baseline.
To address this issue,~\cite{naseer2017deep,wu2017delving,moreau2022lens} leverage additional synthetic training data. Other approaches~\cite{chen2021direct, chen2022dfnet} adapt photometric or feature matching by applying unlabeled data with NeRF synthesis in a semi-supervised manner. However, using unlabeled data from the test set to finetune the APR network is impractical.

Most APR methods focus on improving pose prediction accuracy. However, we demonstrate in the following sections that not all pose predictions are reliable. The APRs mentioned above do not  distinguish accurate predictions from poses with a large error. This paper introduces a modular, plug-and-play method to enable uncertainty estimation for non-UA APRs. Such uncertainty data can be used in the later stages of a visual positioning pipeline to improve tracking robustness and accuracy regardless of the underlying APR.


\subsection{Uncertainty estimation}
Several previous studies have investigated the pose predictions' uncertainty in the training phase of APRs. In Bayesian PoseNet~\cite{kendall2016modelling}, uncertainty is captured by quantifying the variance among multiple inferences of the same input data using Monte Carlo Dropout. Similarly, AD-PoseNet~\cite{huang2019prior} evaluates pose distribution by generating multiple  hypotheses via prior guided dropout. CoordiNet~\cite{moreau2022coordinet}  learns heteroscedastic uncertainty as an auxiliary task during the training. Poses and uncertainties output by CoordiNet are fused into an Extended Kalman Filter (EKF) to smooth the trajectories. Deng \textit{et al.}~\cite{deng2022deep,bui20206d} and Zangeneh \textit{et al.}~\cite{10160466} use pose distributions to represent uncertainty. While these UA APRs offer both pose predictions and uncertainty estimates, the accuracy of estimate poses is much lower than SOTA non-UA APRs that only output poses, as shown in Section~\ref{sec:exp}. Furthermore, existing UA APRs can be time-consuming~\cite{kendall2016modelling,huang2019prior}, require context-specific hyperparameters~\cite{bui20206d}, or display weak extensibility as specific loss functions and modules need to be combined in a training scheme~\cite{bui20206d,10160466,huang2019prior,moreau2022coordinet}.

Our framework enables greater flexibility regarding APR architecture and pose refinement compared to existing UA APRs. Mainstream pre-trained APRs can be integrated into our framework without modifying the network architecture or training schemes. The pose refinement module can leverage the uncertainty data to optimize computations significantly.


%% file: Sections/3Method.tex
\begin{figure}
 \centering
\subfloat[Trajectory (Hospital)]{
 \label{fig:subfig:f} 
 \includegraphics[height=0.8in, width=.3\linewidth]{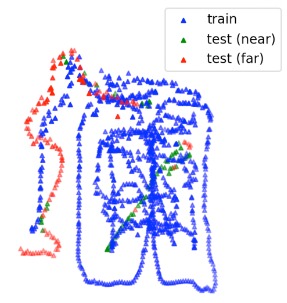}}
  \subfloat[Trans. ]{
 \label{fig:subfig:f} 
 \includegraphics[width=.3\linewidth]{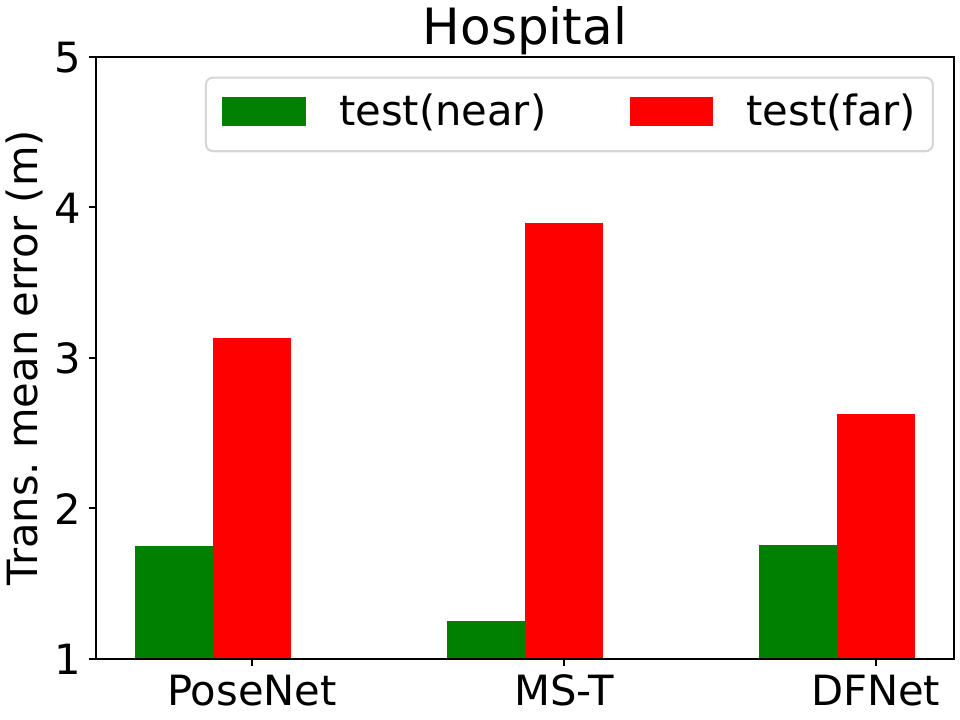}}
\subfloat[Rot. ]{
 \label{fig:subfig:f} 
 \includegraphics[width=.3\linewidth]{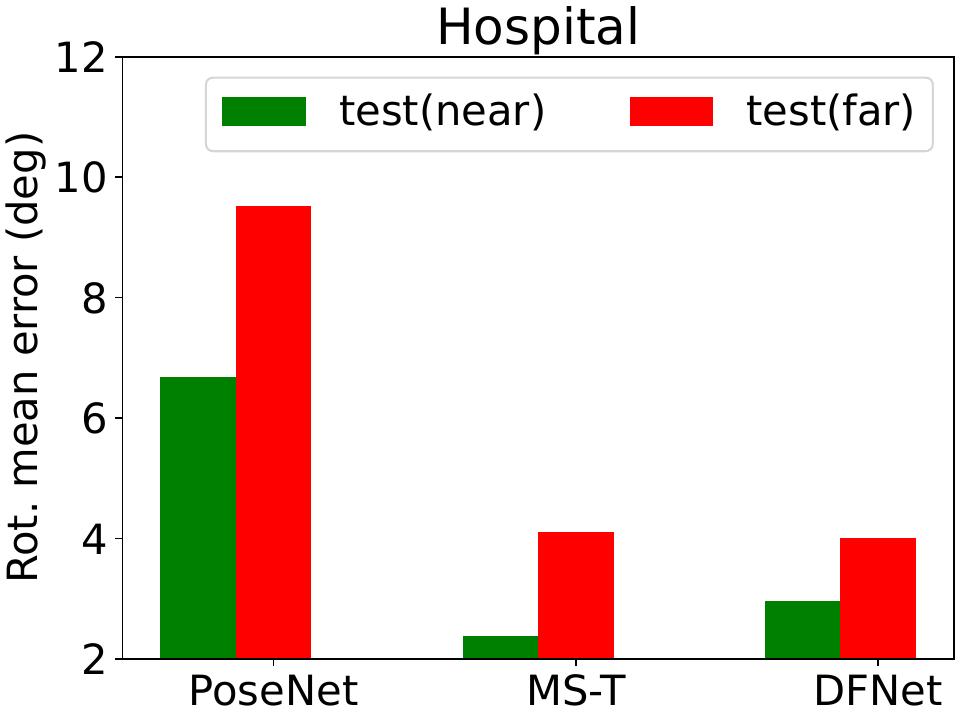}}
 \newline
 \subfloat[Trajectory (Church)]{
 \label{fig:subfig:f} 
 \includegraphics[height=0.8in, width=.3\linewidth]{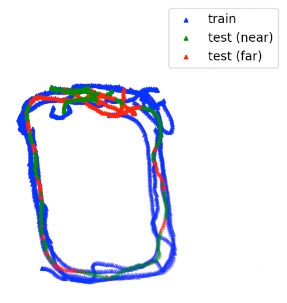}}
  \subfloat[Trans. ]{
 \label{fig:subfig:f} 
 \includegraphics[width=.3\linewidth]{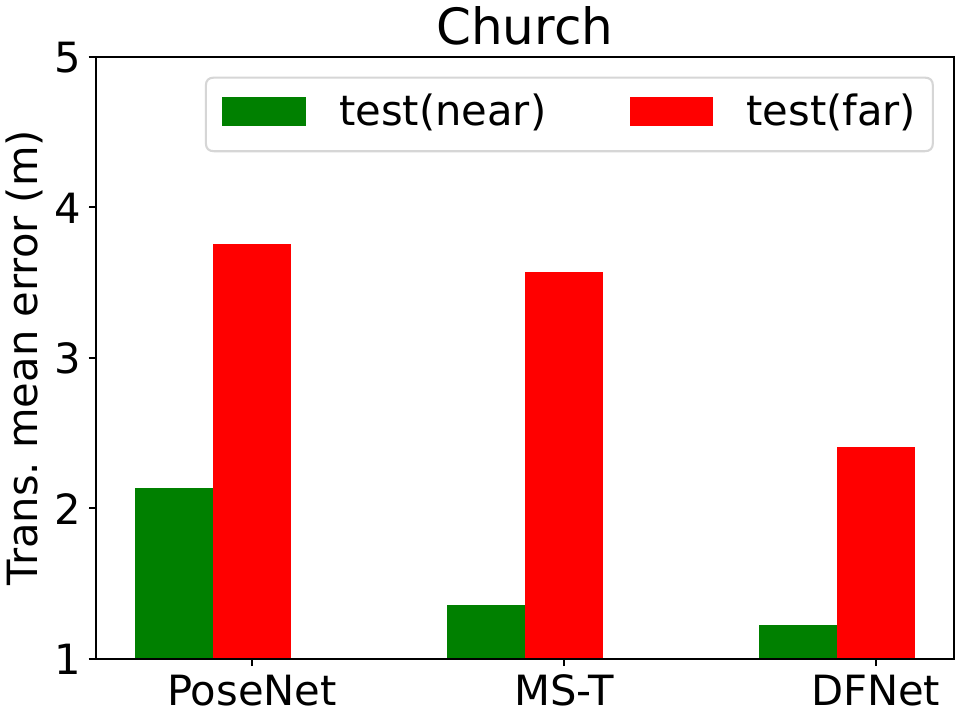}}
\subfloat[Rot. ]{
 \label{fig:subfig:f} 
 \includegraphics[width=.3\linewidth]{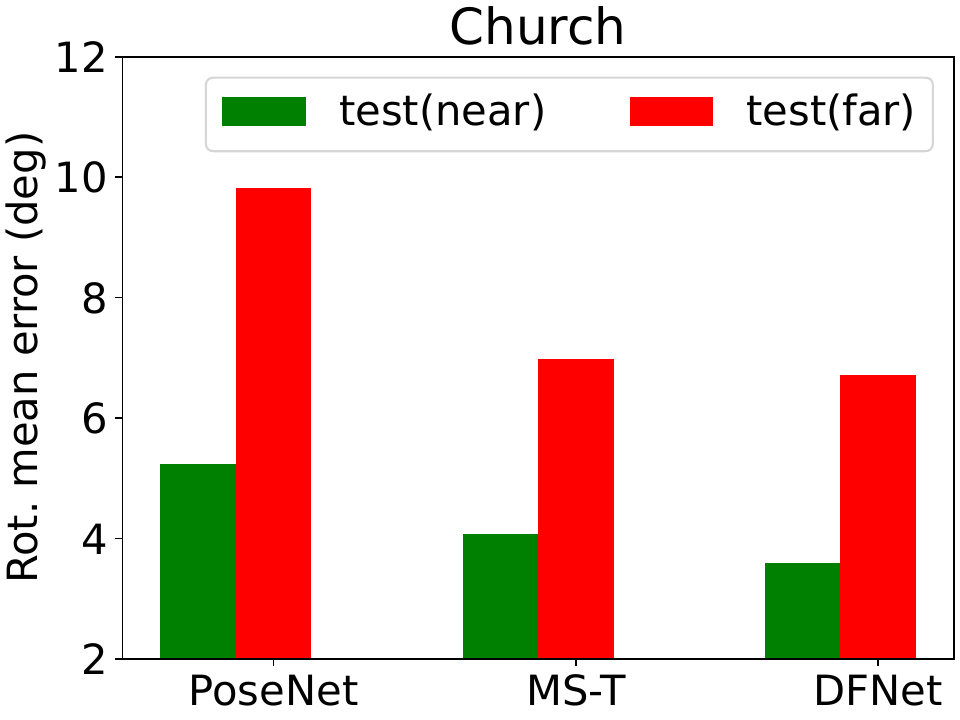}}
 \caption{The accuracy of APR predictions is highly corresponding to pose variants between query and training images. (a) and (d): Camera ground truth trajectory of Hospital and Church in Cambridge Landmarks dataset~\cite{kendall2015posenet}. Blue: training set; Green: queries in the test set near the training set within 2 meters and 10 degrees; Red: queries in the test set far from the training set. (b), (c), (e), and (f) show that all three different APRs~\cite{kendall2015posenet,shavit2021learning,chen2022dfnet} 
 have better predictions on test (near) than test (far).}
 \label{fig:track} 
\end{figure}

\section{METHOD}
\label{sec:method}
Our objective is to enhance the robustness of APR methods and reduce the refinement overhead by identifying the reliabilities of predictions. Most existing APR methods suffer from limited generalizability due to overfitting their training sets.  We first investigate several prevalent APR methods to show the accuracy of APR predictions is highly corresponding to pose variants between query and training images. The result is illustrated in Figure~\ref{fig:track}, where all three APR models exhibit more accurate predictions for test queries with viewpoints similar to those in the training set, compared to queries that fall outside the coverage of the training data. These quantitative results are also validated by the similar observation of ~\cite{sattler2019understanding,ng2021reassessing}.
 In light of this, we believe that the proximity of viewpoints also implies that these images have similar features. Therefore, we exploit pose predictions of APR to obtain the most corresponding features from the database. We then model the uncertainty of predictions by measuring the similarity between the features of the query image and the corresponding features of the database. The uncertainty information is further used as pose refinement constraints which can reduce the computational cost while achieving comparable performance.

\subsection{Uncertainty estimation module}
\begin{figure}[!h]
  \centering
  \includegraphics[width=0.48\textwidth]{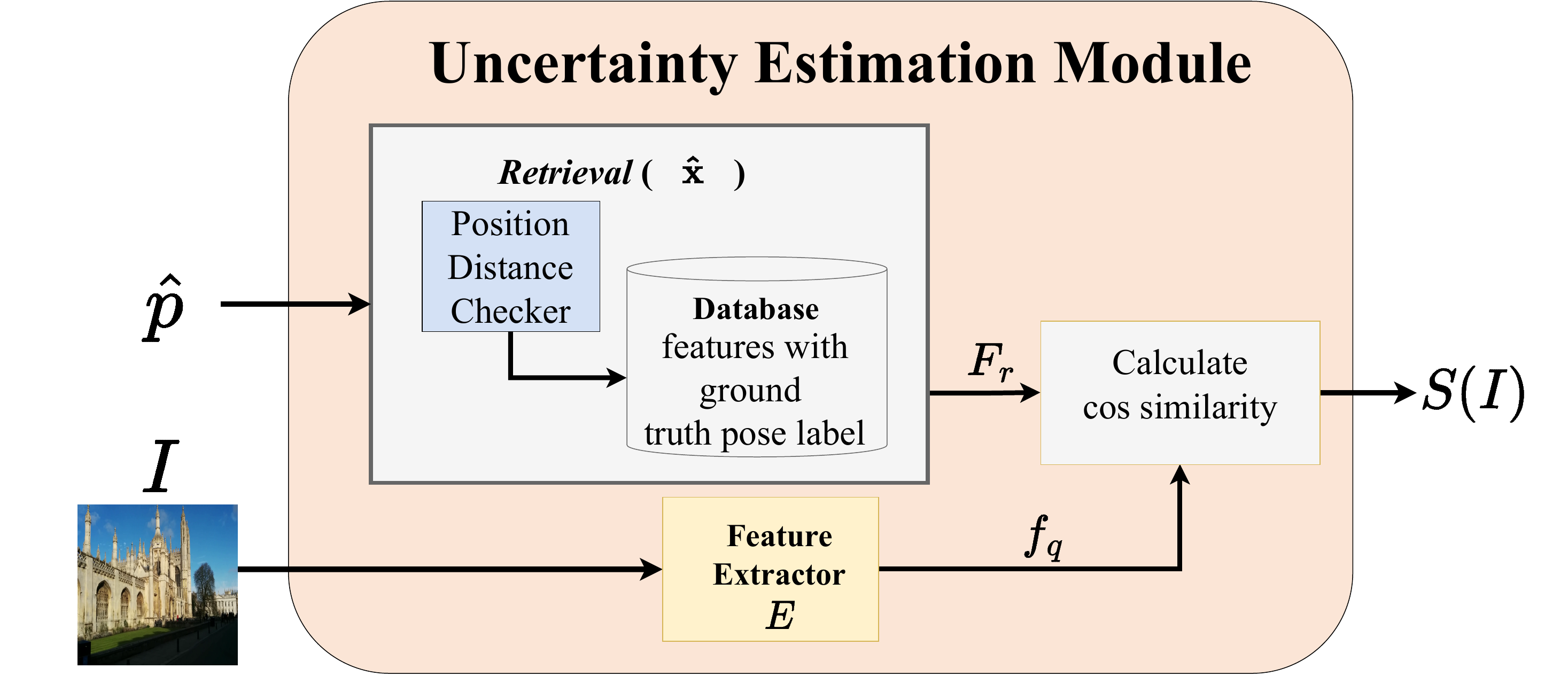}
  \caption{Uncertainty estimation module.}
\label{fig:uemodule}
\end{figure}

 The uncertainty estimation module of \sysname is shown in Figure~\ref{fig:uemodule}. We build a database that stores low-dimensional feature embeddings of images from the training set and their associated ground truth poses, which operates as follows:
\begin{enumerate}
     \item  Given a query image $I$, APR $P$ outputs estimated translation $\mathbf{\hat{x}}$ and rotation $\mathbf{\hat{q}}$ so that $P(I) = \hat{p} = <\mathbf{\hat{x}},\mathbf{\hat{q}}>$. All feature embeddings $F_r$ with poses within ranges of threshold $d_{th}$ to $\mathbf{\hat{x}}$ are retrieved. Then, proceed to step 2). If there is no valid pose in the database, we simply assume that the similarity is 0 and skip to step 3).
     
    \item A feature extractor $E$ extracts the feature embeddings of the query image $I$ as $f_q$. Then, we compute the cosine similarity between $f_q$ and each feature embedding $f_r^i \in F_r,$ retrieved in step 1). We take the maximum value as the final similarity score.
    
    \item  The estimated pose that obtain a similarity score above threshold $\gamma$ is deemed to be reliable, and thus to be output directly or refined with small steps. Otherwise, the poses are considered unreliable and to be treated with more refinement steps.
\end{enumerate}

Compared to typical image retrieval approaches that often require large computational complexity in the visual feature descriptor matchings, our pose retrieval is a super cheap alternative by only relying on 3-D position searches with a significant reduction in the search space. The database consisting of pose-feature embedding pairs also saves a lot of memory footprint compared to the image database in ~\cite{sarlin2019coarse}.

\subsection{CNN feature embeddings}


Our feature extractor $E$ is similar to an ordinary PoseNet~\cite{kendall2015posenet,kendall2017geometric}, which predicts a 6-DoF camera pose for an input image.   $E$ has an FC layer before the final regressor of feature size $n = 1024$.  The difference is that $E$ also outputs the FC layer as feature embedding.
$E$ is supervised by the learnable loss function~\cite{kendall2017geometric}:
\begin{align}
\mathcal{L}_\delta &= \mathcal{L}_x\exp(-\hat{s}_x) + \hat{s}_x + \mathcal{L}_q \exp(-\hat{s}_q) + \hat{s}_q
\label{eq:deltaxq}
\end{align}
, where $\mathcal{L}_x = ||\mathbf{\hat{x}}-\mathbf{x}||_2$ and $\mathcal{L}_q  = ||\frac{\mathbf{\hat{q}}}{||\mathbf{\hat{q}}||}-\mathbf{q}||_2$. To measure the difference between two feature embeddings, we
compute cosine similarity between feature embedding $f_{q}$ extracted from query images and each retrived feature embeddings $f_{r}^{i}, f_{r}^{i}\in F_r$.
\begin{equation}
    \cos(f_{q},f_{r}^{i} ) = \frac{f_{q} \cdot f_{r}^{i} }{||f_{q}||_2\cdot||f_{r}^{i}||_2}
\end{equation}
, where $f_q, f_{r}^{i} \in \mathrm{R}^{1\times 1024}$. The Similarity score of the query image $I$ is:
\begin{equation}
    S(I) = \max(\cos(f_{q},f_{r}^{i} )), f_{r}^{i}\in F_r.
\end{equation}
The similarity score of the query image $I$ is $S(I)$, where $-1\leq S(I) \leq 1$, which reflects the reliability of the prediction.


\subsection{Outcomes}
\label{subsec:outcomes}
The proposed uncertainty estimation module can identify an unreliable pose $\hat{p}$ under two scenarios:
\begin{enumerate}
    \item The training set does not contain an image close enough to the estimated location $\mathbf{\hat{x}}$.
    \item Although feature embeddings of images in training set are retrieved from the database based on $\mathbf{\hat{x}}$, they all present too few similar features. 
\end{enumerate}
In the first scenario, $\hat{p}$ can only get 0 similarity score because of the limited generalization ability of the neural network. 
The APR thus has a high chance of predicting a vastly incorrect pose for an image $I$ far from the training set. 
In the second scenario, although valid most-similar feature embeddings $F_r$ extracted from images in the training set is found based on $\mathbf{\hat{x}}$, all $f_r^i\in F_r$ are not similar enough to $f_q$. Either the orientations of the query image and these images are very different, leading to little overlap, or the predicted  $\mathbf{\hat{x}}$ and $\mathbf{\hat{q}}$ have a large error, and $F_r$ found according to this pose is a false positive.
\subsection{Pose refinement module}
In this paper, we incorporate the test-time Neural feature synthesizer (NeFeS) refinement pipeline~\cite{chen2023refinement} in our pose refinement module. We modify the refinement procedure to be uncertainty-aware.  The refinement procedure is as follows:  (i) For estimated camera pose $\hat{p}$ with a similarity score in the uncertainty estimation module, NeFeS $N$ renders a dense feature map $m^{rend}$ given $\hat{p}$. (ii) At the same time, a feature map extractor $G$ extracts a dense feature map $m^G = G(I)$ from the query image. (iii) The pose $\hat{p}$ is iteratively refined by minimizing the feature cosine similarity loss~\cite{chen2023refinement} between $m^{rend}$ and $m^G$. The refined steps depend on the similarity score $S(I)$ in our \textit{uncertainty estimation module}.

\subsection{Hyperparameters}
\label{sec:hyper}
During test time, the proposed method is controlled by two hyperparameters: the distance threshold $d_{th}$, and the similarity threshold $\gamma$. $d_{th}$ depends on the size of the scene.

%% file: Sections/4Experiment.tex
\section{EXPERIMENT}

\begin{figure*}[h]
 \centering
  \subfloat[Trans. (DFNet$^{hr}$ + Filter)]{
 \label{fig:subfig:f} 
 \includegraphics[width=0.32\linewidth]{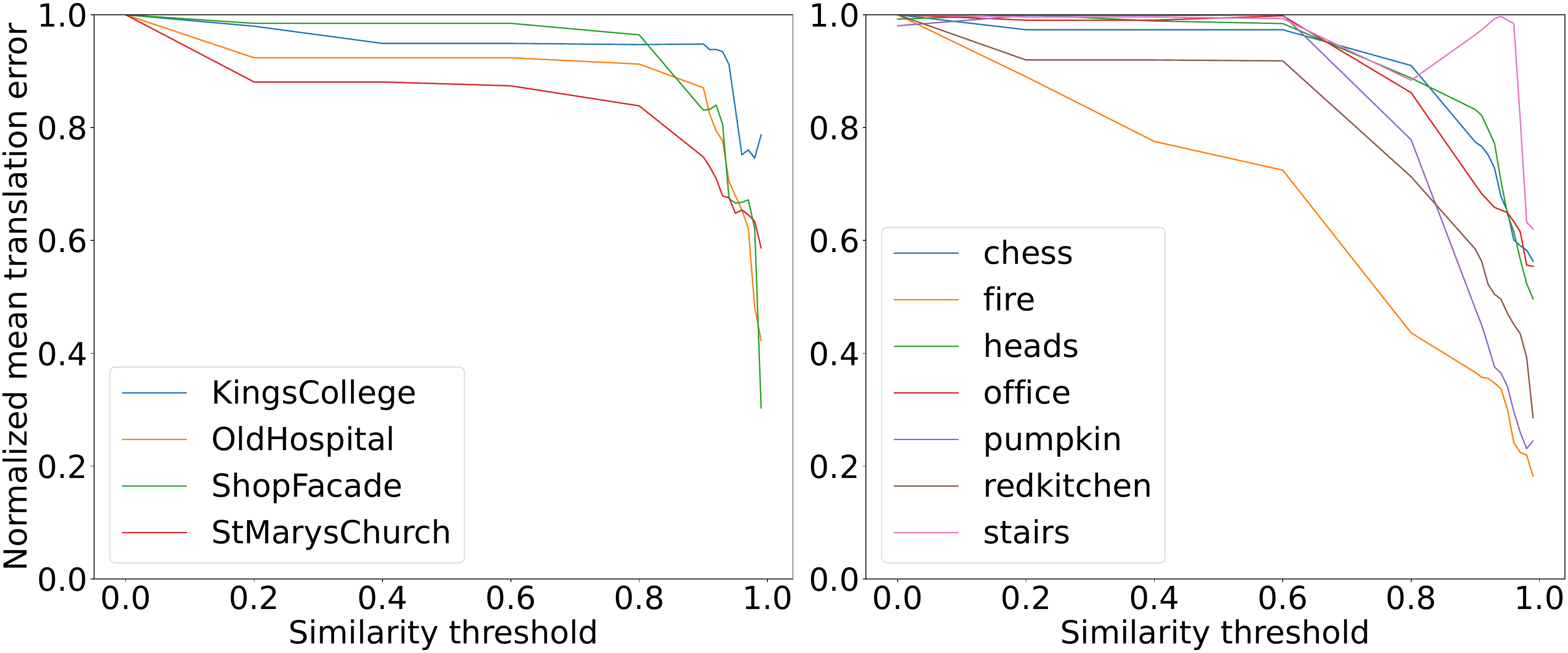}}
 \subfloat[Trans. (MS-T$^{hr}$ + Filter)]{
 \label{fig:subfig:f} 
 \includegraphics[width=0.32\linewidth]{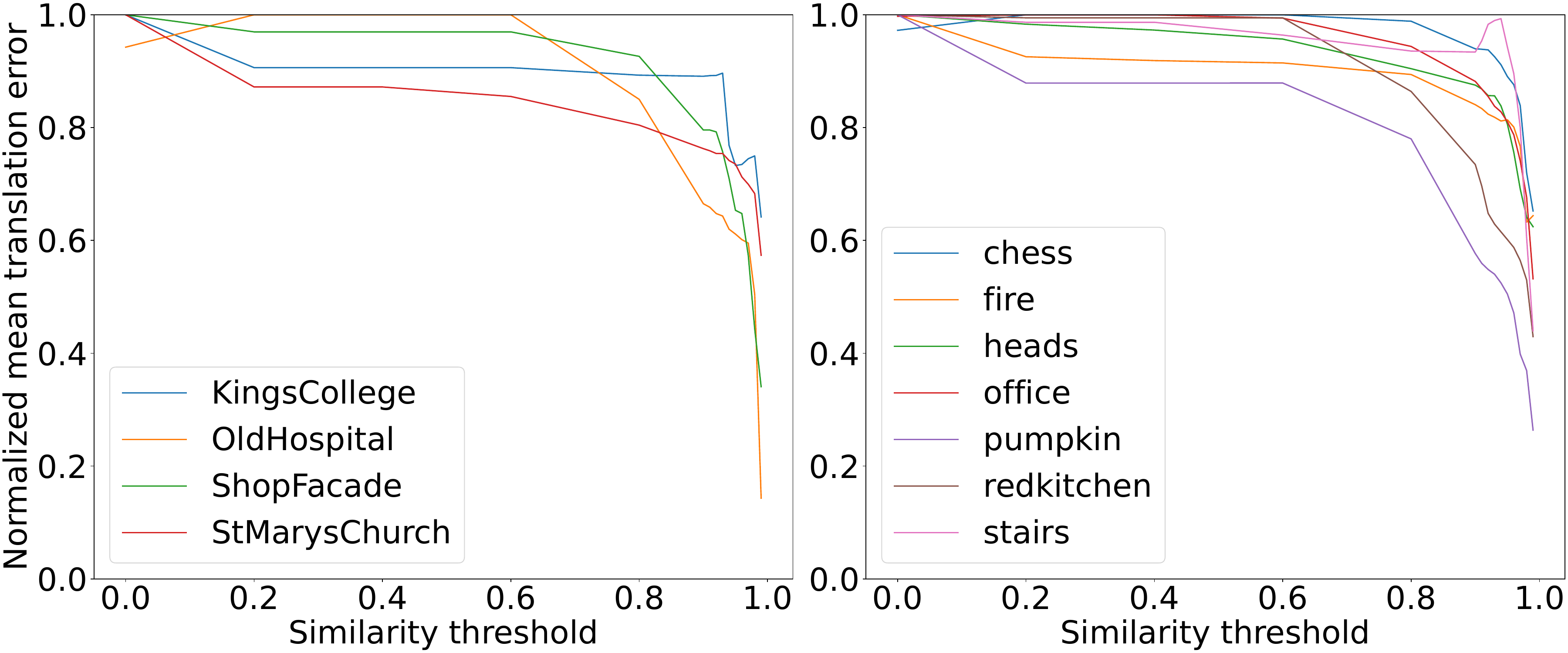}}
 \subfloat[Trans. (PN$^{hr}$ + Filter)]{
 \label{fig:subfig:a} 
 \includegraphics[width=0.32\linewidth]{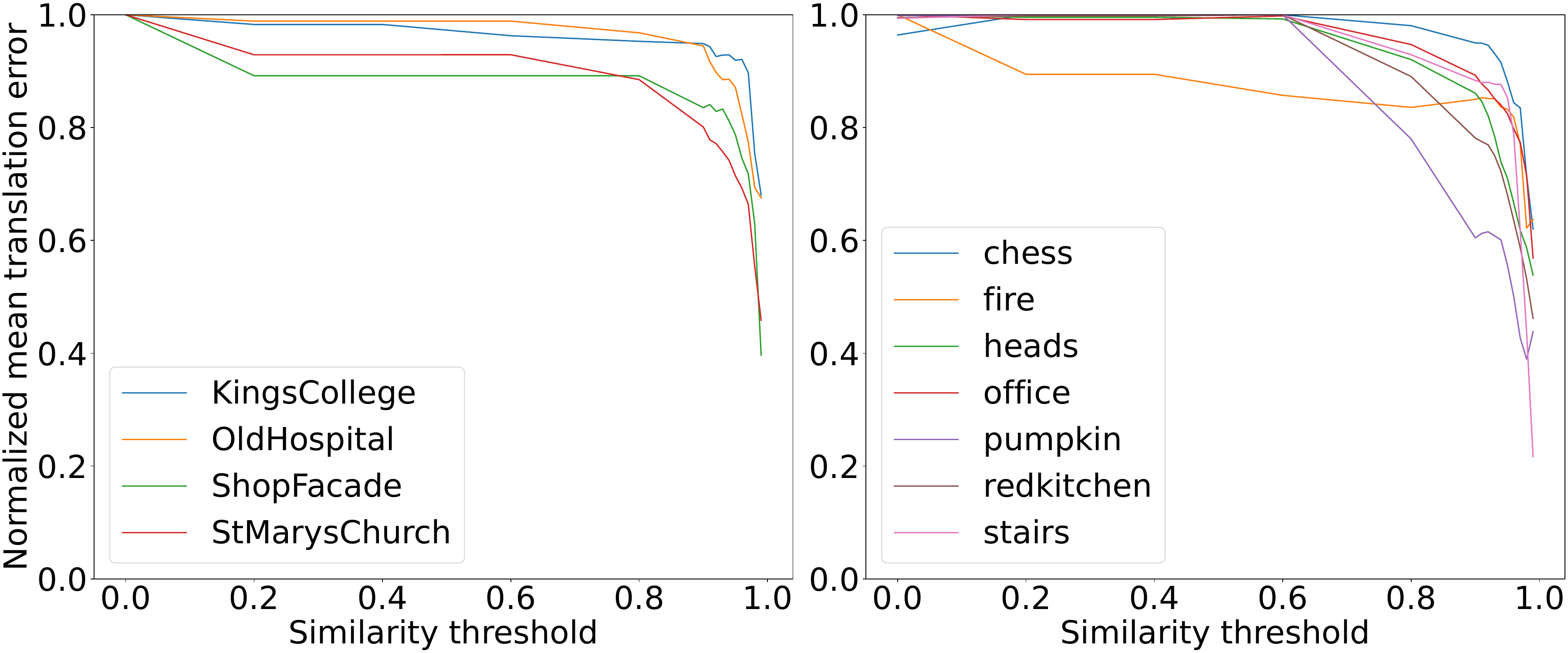}}
 \newline
 \subfloat[Rot. (DFNet$^{hr}$ + Filter)]{
 \label{fig:subfig:a} 
 \includegraphics[width=0.32\linewidth]{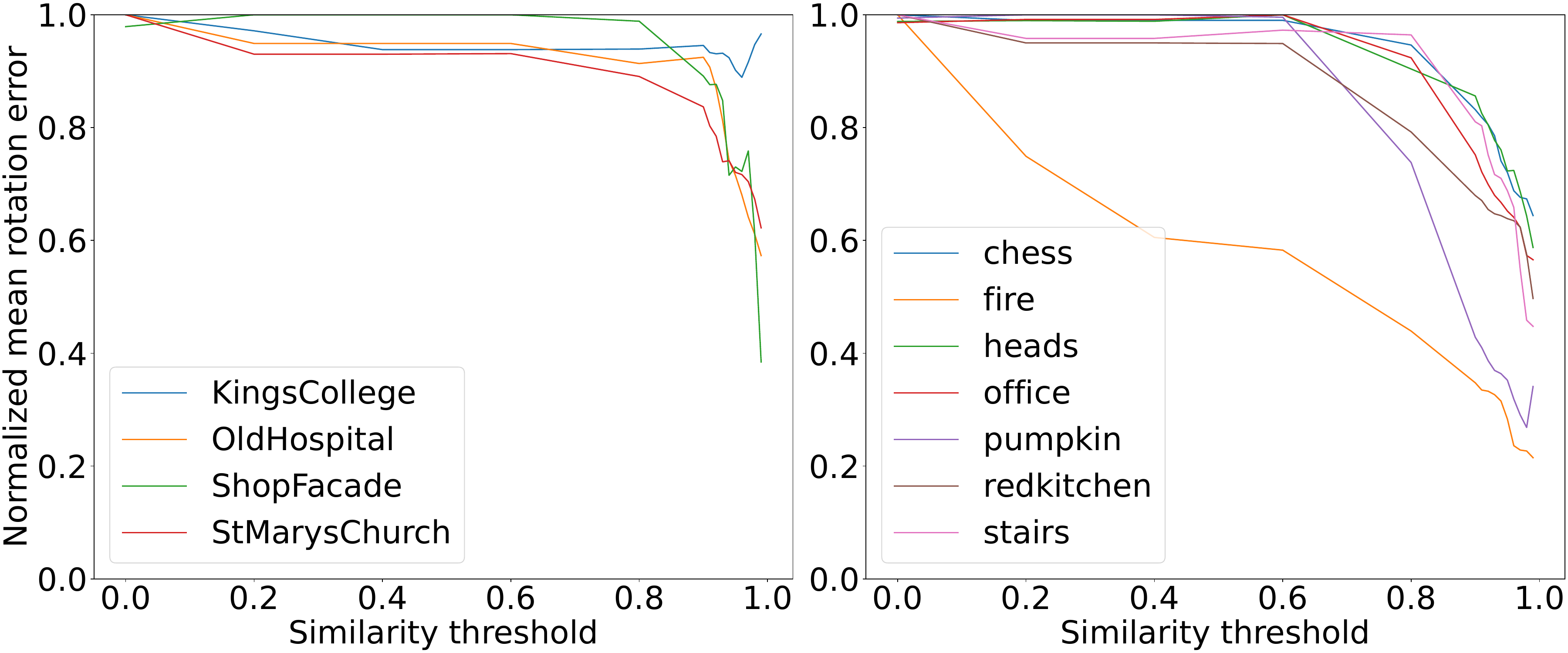}}
 \subfloat[Rot. (MS-T$^{hr}$ + Filter)]{
 \label{fig:subfig:a} 
 \includegraphics[width=0.32\linewidth]{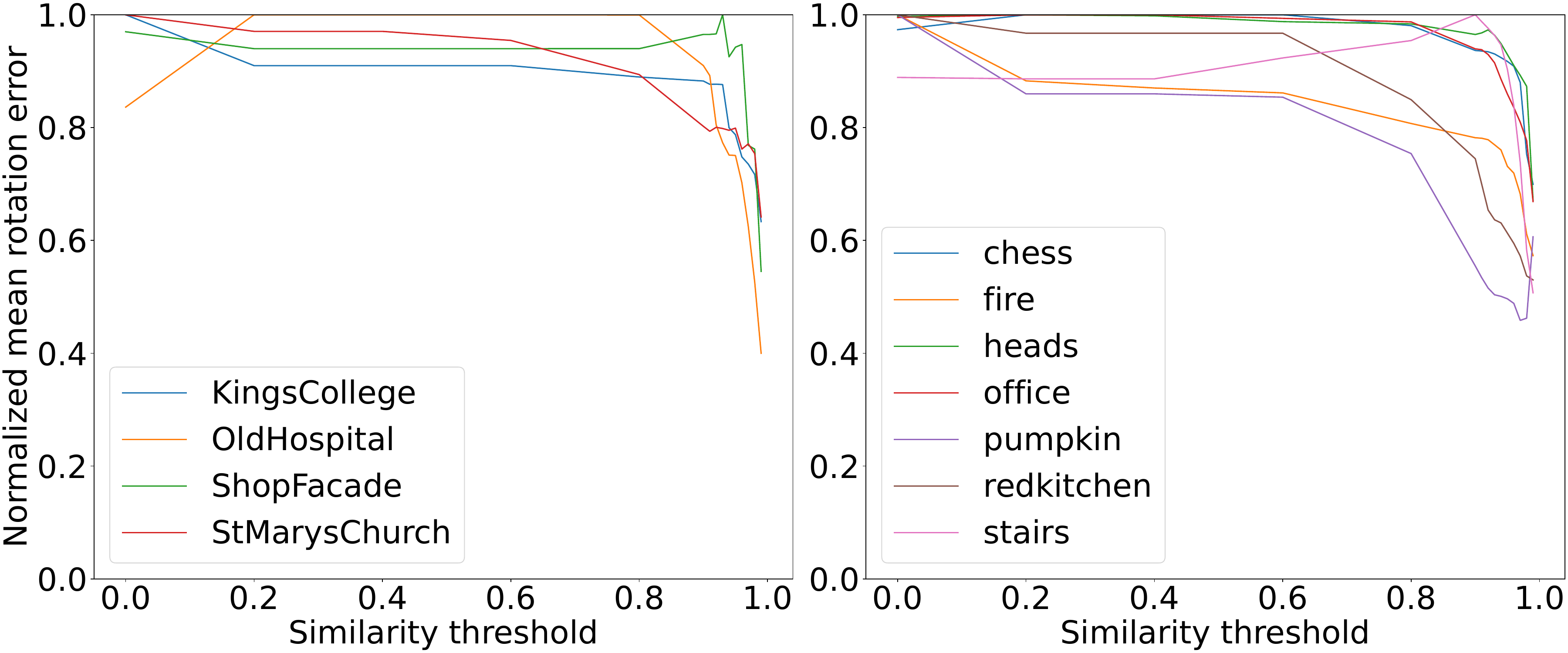}}
  \subfloat[Rot. (PN$^{hr}$ + Filter)]{
 \label{fig:subfig:a} 
 \includegraphics[width=0.32\linewidth]{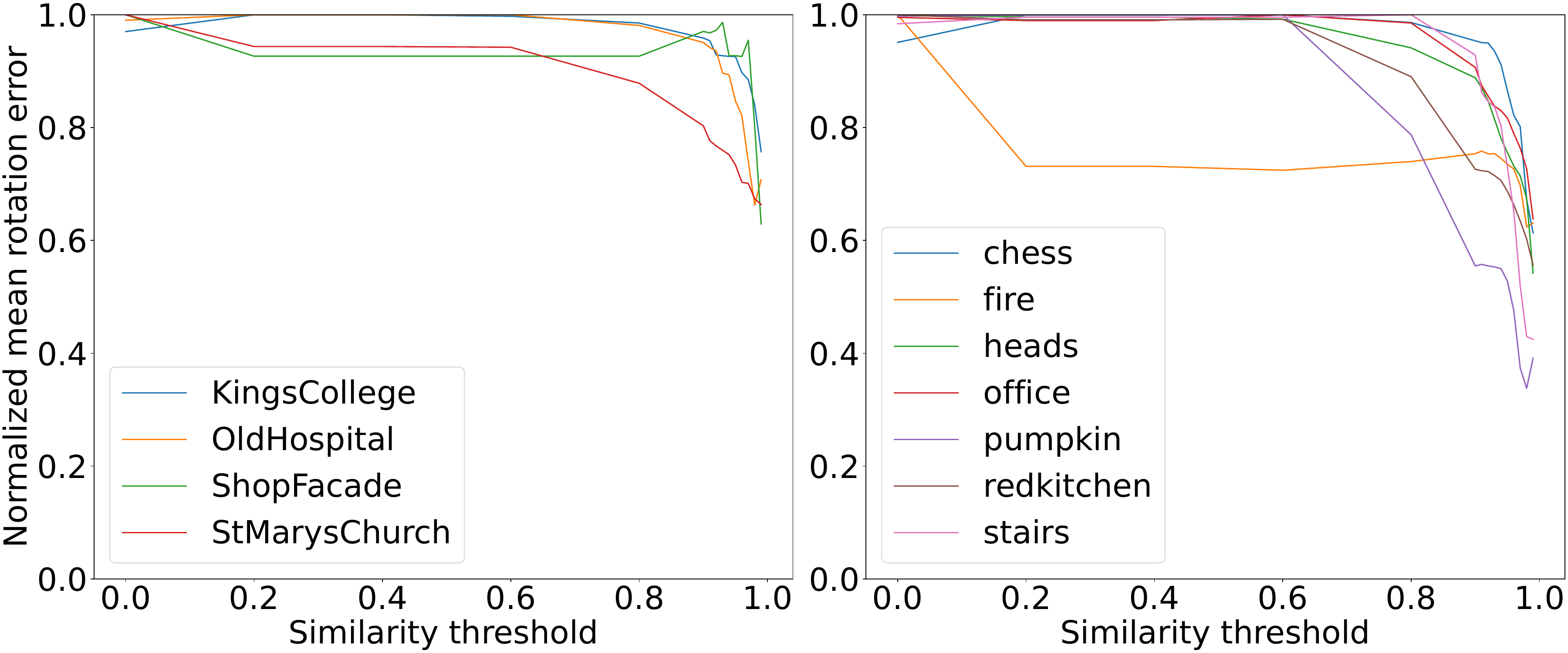}}
 \newline
  \subfloat[Retained Predictions (\%) (DFNet$^{hr}$ + Filter)]{
 \label{fig:subfig:a} 
 \includegraphics[width=0.32\linewidth]{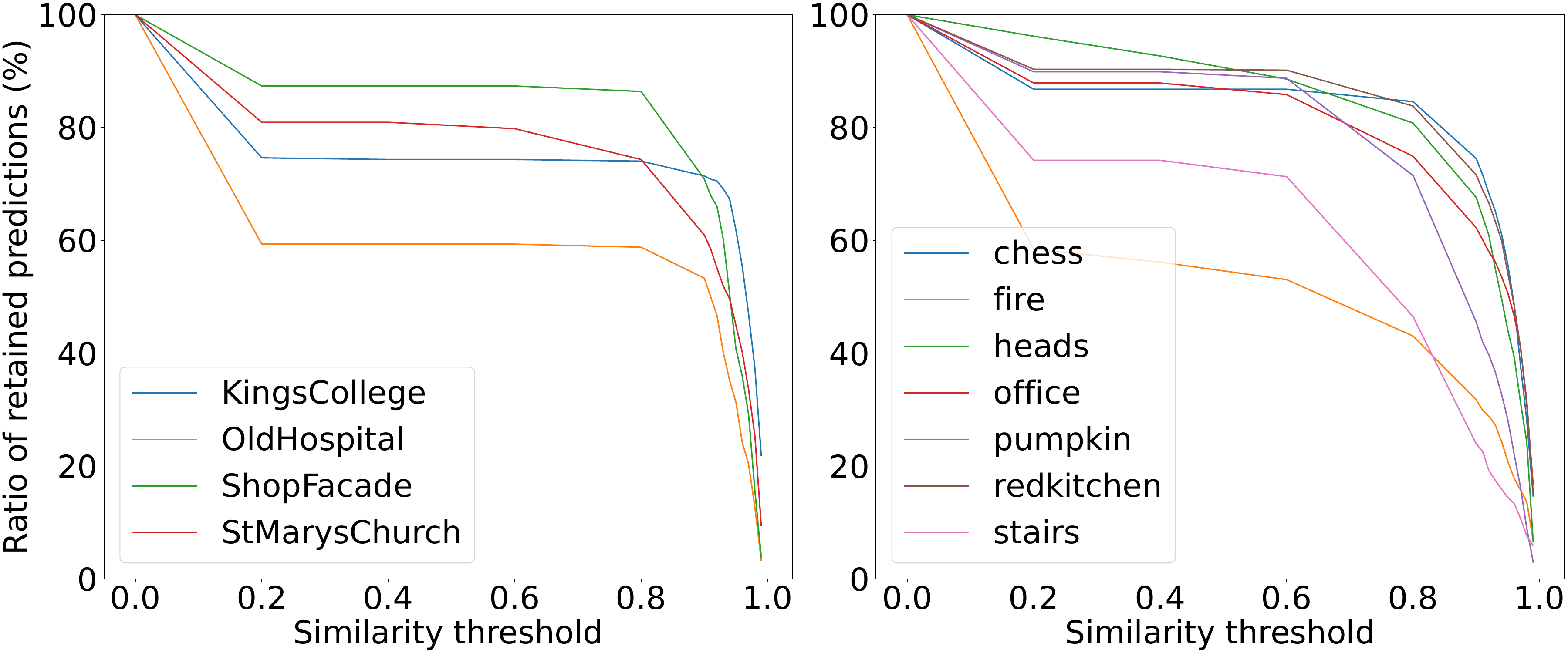}}
 \subfloat[Retained Predictions (\%) (MS-T$^{hr}$ + Filter)]{
 \label{fig:subfig:a} 
 \includegraphics[width=0.32\linewidth]{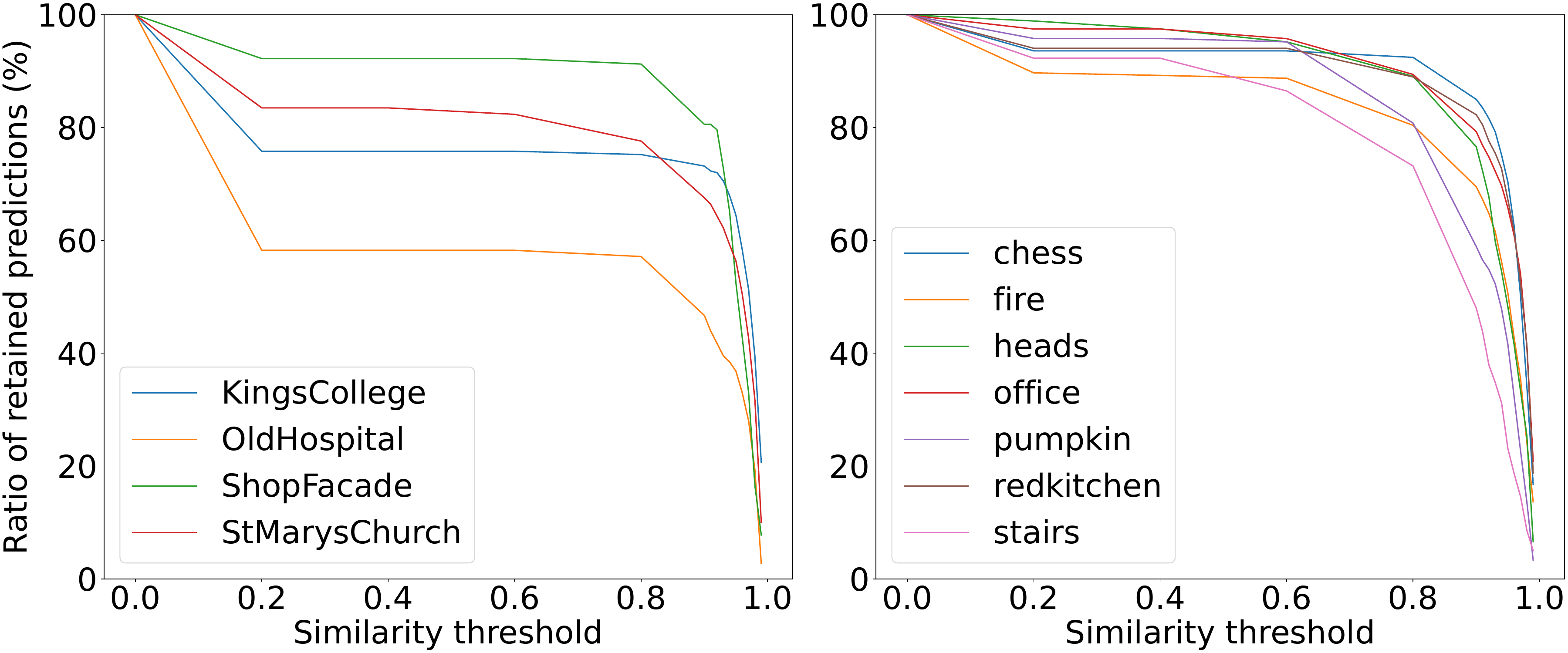}}
  \subfloat[Retained Predictions (\%) (PN$^{hr}$ + Filter)]{
 \label{fig:subfig:a} 
 \includegraphics[width=0.32\linewidth]{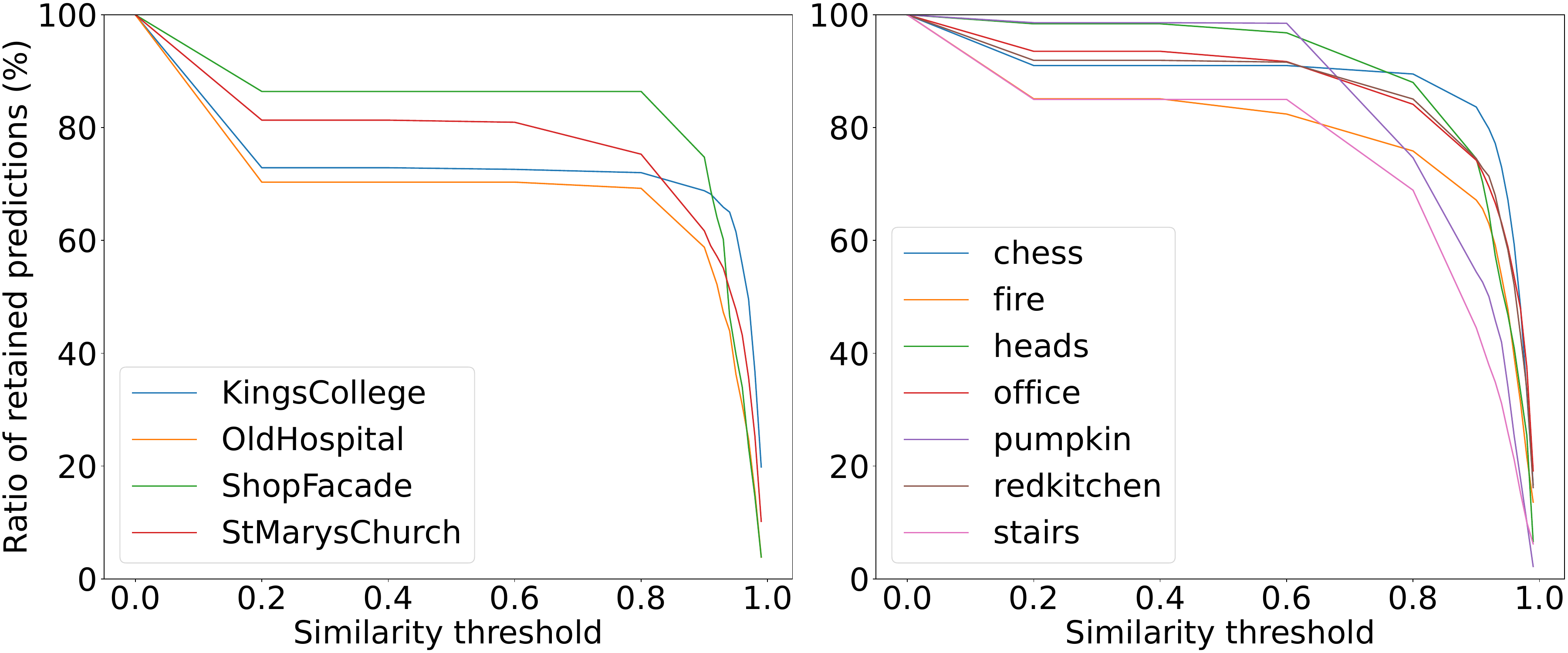}}
 \caption{Uncertainty evaluation on the 7Scenes and Cambridge Landmarks datasets. Subfigures (a)-(f) show the correlation between similarity threshold $\gamma$ and normalized pose error (0-1). Based on the similarity threshold, uncertain samples with low similarity scores are gradually removed. Subfigures (g)-(i) in the last row show the ratio of retained predictions (\%) as threshold increases.
 We observe that as we remove the samples with low similarity scores the overall error drops indicating a clear correlation between our predictions and the actual inaccurate predictions.}
 \label{fig:correlation} 
\end{figure*}
\label{sec:exp}
To demonstrate the scalability and effectiveness of \sysname, we integrate the latest representative mainstream APR architectures into our framework and test them over indoor and outdoor datasets of various scales.

\subsection{Datasets}
The 7Scenes dataset~\cite{glocker2013real,shotton2013scene} is an indoor dataset consisting of seven small scenes from $1m^3$
to $18m^3$. Each scene contains a training set with 1000 to 7000 images and a test set with 1000 to 5000 images.  We
use the Structure from Motion (SfM) ground truth provided by~\cite{brachmann2021limits} for our experiment. The Cambridge Landmarks~\cite{kendall2015posenet} dataset represents six large-scale outdoor scenes ranging from $900m^2$ to $5500m^2$. We utilize four out of six scenes for comparative evaluation, and the SfM ground truth provided by~\cite{kendall2015posenet}. Each scene contains 231 to 1487 images for training and 103 to 530 for testing.

\subsection{Implementation Details}
In this paper, we train the feature extractor $E$ with EfficientNet-B0~\cite{tan2019efficientnet} as the backbone in the \textit{uncertainty estimation module}. The training process of $E$ follows~\cite{kendall2015posenet,kendall2017geometric}.  It is trained using Adam optimizer~\cite{kingma2014adam} with an initial learning rate of $1e^{-4}$ and a weight decay of $5e^{-4}$. During training, all input images are resized to $256 \times 256$ and then
randomly cropped to $224 \times 224$. 

We implement the \textit{uncertainty estimation module} of \sysname over three recent APR models: \textbf{PoseNet (PN)}~\cite{kendall2015posenet} is the
classic pose regression architecture. \textbf{ MS-Transformer (MS-T)}~\cite{shavit2021learning}  aggregates the activation maps with self-attention and queries the scene-specific information. \textbf{DFNet}~\cite{chen2022dfnet} is trained by feature matching loss. We train these three models on our platform using the open-source codes.
\textbf{We  incorporate these APRs into our \sysname framework, referring to them as $\text{APR}^{hr}$}.   We set $d_{th}$ as $0.2m$ for the 7Scenes dataset and $1.5m$ for the Cambridge dataset. We set $\gamma$ range from $0.95\sim0.98$. Furthermore, we show the impact of $\gamma$ in the next subsection.

In this paper, we implement the NeFeS refinement pipeline~\cite{chen2023refinement} in our pose refinement module. 
The term $\text{APR}^{hr} + \text{NeFeS}^{hs(x)}_{ls(y)}$ denotes the refinement process based on the similarity of estimated poses. We refine high similarity (hs) poses that are above the similarity threshold $\gamma$ with $x$ steps. Conversely, we refine low similarity (ls) poses that are below the similarity threshold $\gamma$ with $y$ steps. For instance, $\text{APR}^{hr} + \text{NeFeS}^{hs10}_{ls50}$ indicates that we refine poses with similarity scores higher than $\gamma$ for 10 steps, but refine poses with a low similarity (ls) $< \gamma$ for 50 steps. Alternatively, we use $\text{APR}^{hr} + \text{Filter}$ to denote the scenario where we exclude the pose refinement module and directly filter out estimated poses with low similarity scores (ls). This approach helps us analyze the effectiveness of our framework, as shown in the uncertainty evaluation below. Besides, $\text{APR}^{hr} + \text{Filter}$ can be used to reject bad predictions in real mobile robotics applications simply.

\subsection{Uncertainty Evaluation}

\begin{figure}[h]
 \centering
\subfloat[Fire (DFNet$^{hr}$)]{
 \label{fig:subfig:f} 
 \includegraphics[width=.3\linewidth]{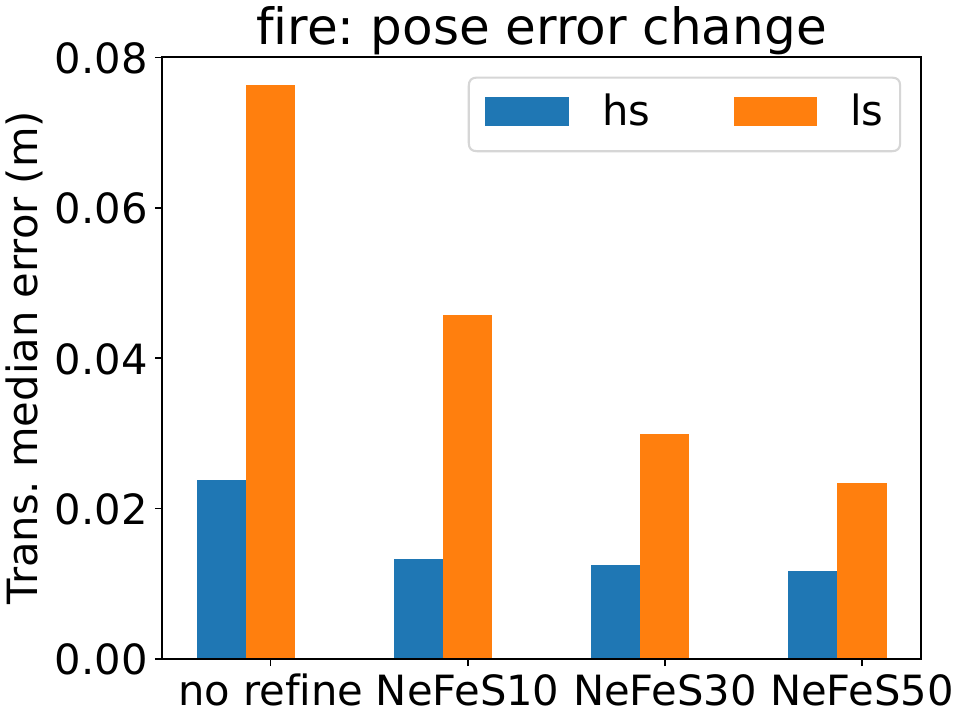}}
  \subfloat[Heads (DFNet$^{hr}$)]{
 \label{fig:subfig:f} 
 \includegraphics[width=.3\linewidth]{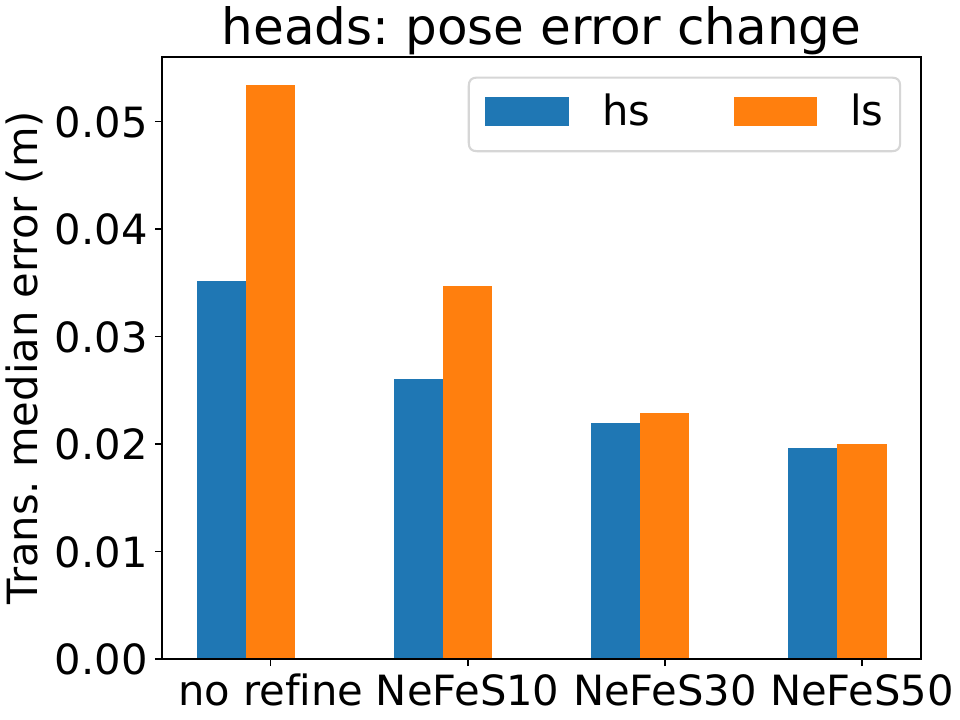}}
     \subfloat[Pumpkin (DFNet$^{hr}$)]{
 \label{fig:subfig:f} 
 \includegraphics[width=.3\linewidth]{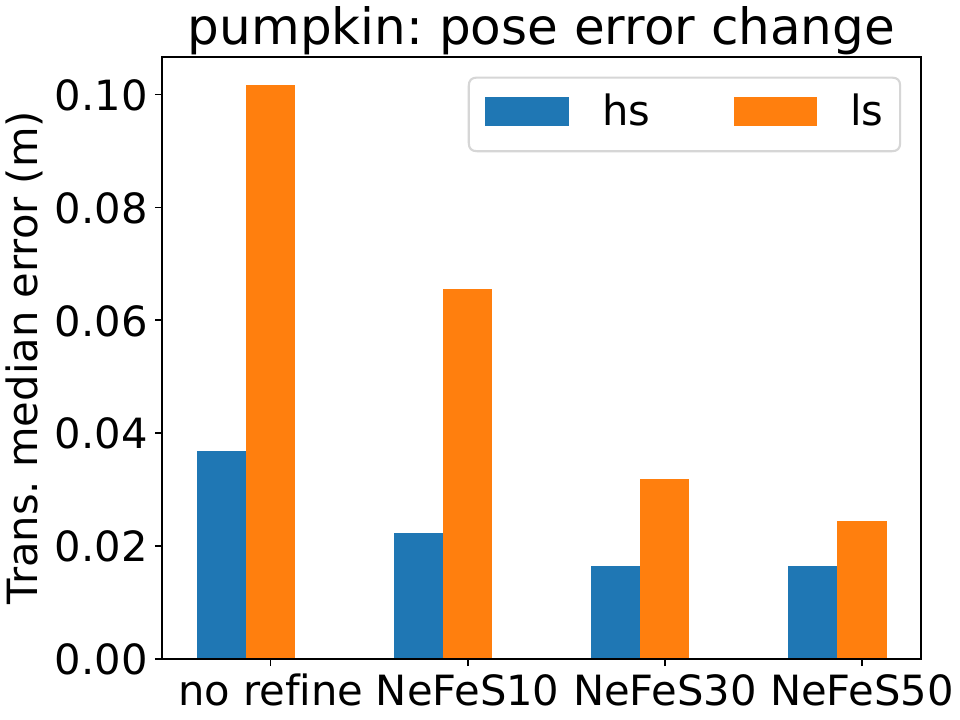}}
 \newline
 \subfloat[Fire (DFNet$^{hr}$)]{
 \label{fig:subfig:a} 
 \includegraphics[width=.3\linewidth]{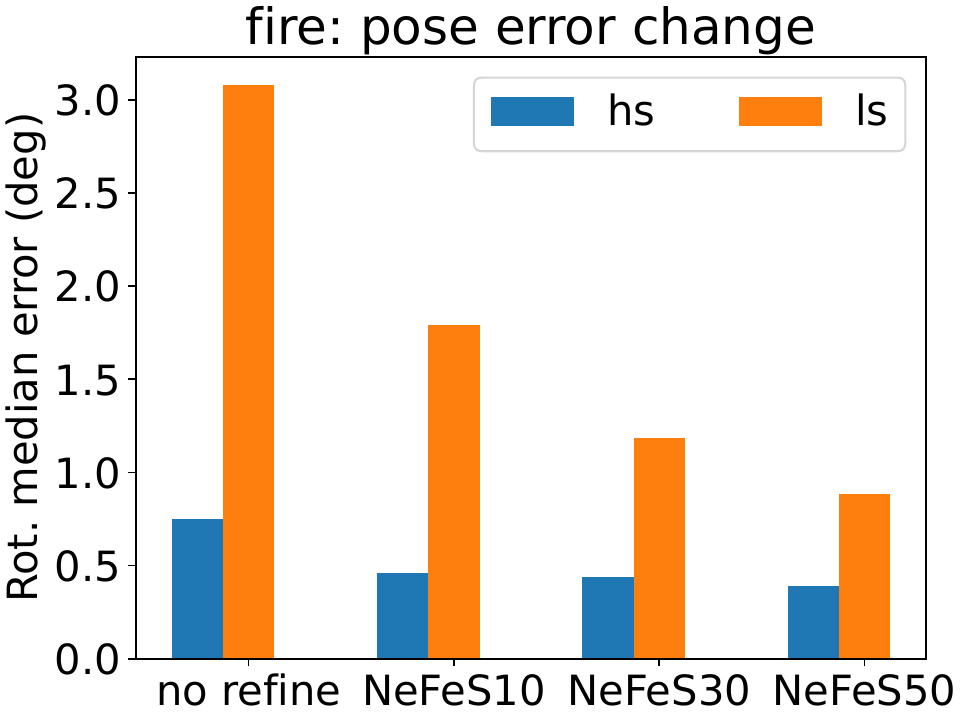}}
 \subfloat[Heads (DFNet$^{hr}$)]{
 \label{fig:subfig:a} 
 \includegraphics[width=.3\linewidth]{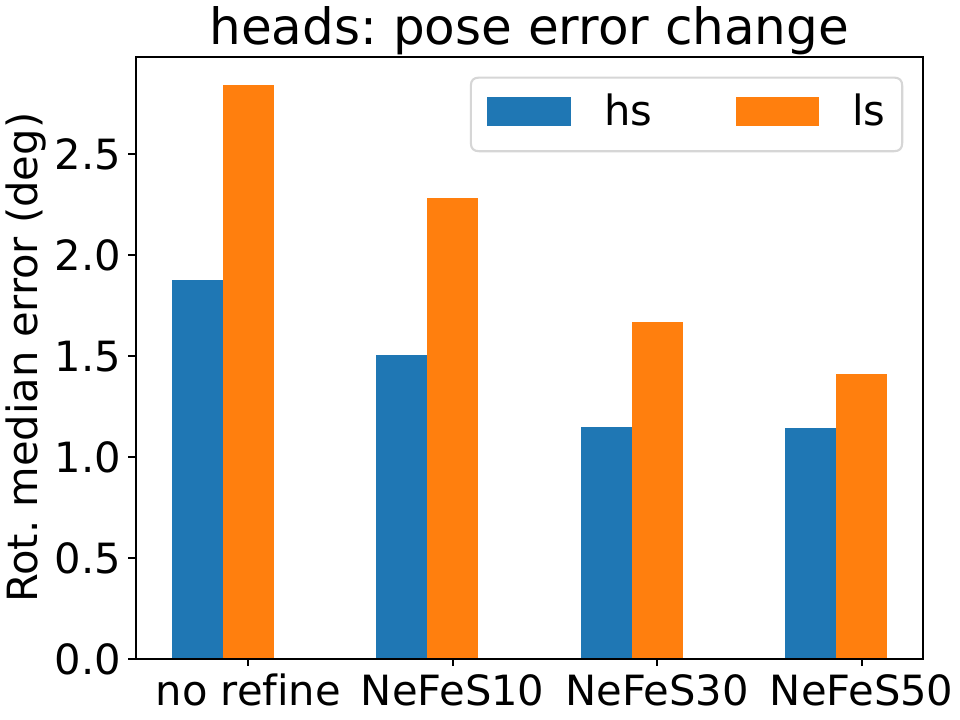}}
 \subfloat[Pumpkin (DFNet$^{hr}$)]{
 \label{fig:subfig:a} 
 \includegraphics[width=.3\linewidth]{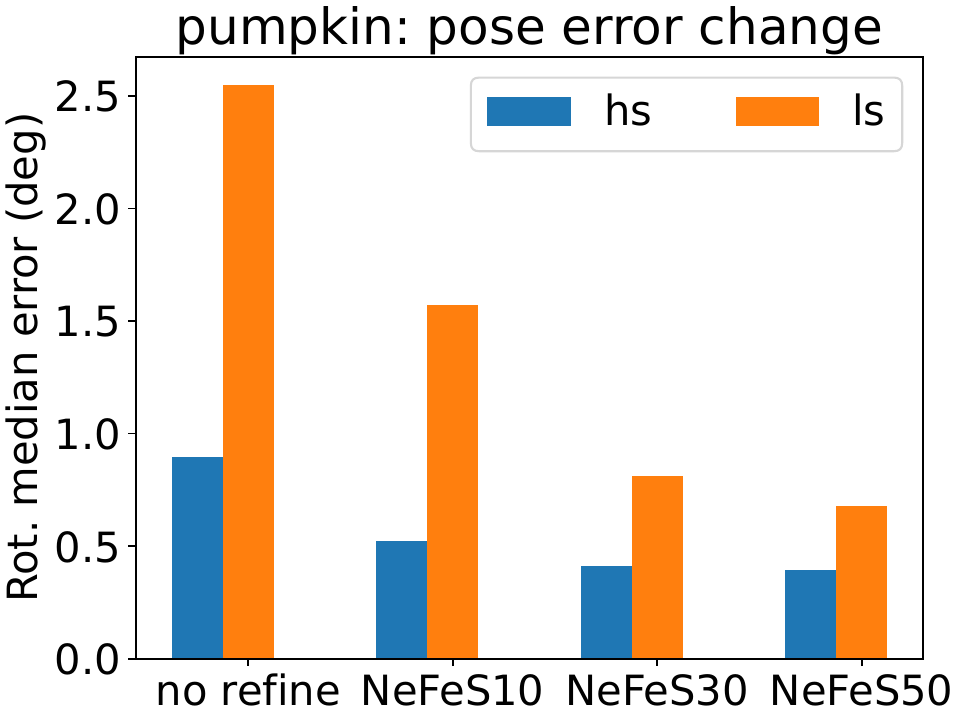}}
 \caption{Plots of translation and rotation errors against the
number of iteration for images pass the similarity threshold (hs) and images with low similarity scores (ls) do not pass the similarity threshold on some scenes of 7Scenes. The hs predictions ratio of each scene is provided in Table~\ref{tab:df_level}. NeFeS$m$ denotes runing the refinement process for $m$ iterations. }
 \label{fig:bar_7s} 
\end{figure}

\begin{table}[h]
\caption{Percentage of predictions with high (0.25m, $2^{\circ}$), medium (0.5m, $5^{\circ}$), and low (5m, $10^{\circ}$) accuracy~\cite{sattler2018benchmarking} (higher is better). The value in parentheses represents the ratio of retained predictions (hs) identified in the test set by our method. }
\centering
$$
\begin{tabular}{l|crr}
\hline
&Dataset   &  DFNet \cite{chen2022dfnet}    & DFNet$^{hr}$ + Filter (ours)   \\ \hline
7Scenes &Chess & 81.1/97.3/\textbf{100} &\textbf{87.6/99.6/100} (75\%) \\
&Fire&46.5/75.7/94.5 &\textbf{92.3/97.1/100} (21\%) \\
&Heads & 44.8/87.2/98.1&\textbf{56.6/98.4/100} (44\%) \\
&Office &63.6/89.7/96.6 &  \textbf{82.7/97.9/99.9} (31\%) \\ 
&Pumpkin&51.6/77.6/93.1 &\textbf{80.5/99.1/100} (28\%)  \\
&Kitchen&59/82.1/94.5& \textbf{77.9/99.6/100} (30\%) \\
&Stairs &32.9/79.7/98.5 & \textbf{66/100/100} (14\%)  \\\hline
Cambridge &Kings &  4.1/27.1/96.5& \textbf{5.3/28.9/100} (55\%) \\
&Hospital& \textbf{0}/1/94.5&\textbf{0/4.3/100} (13\%) \\
&Shop &7.8/31.1/97.1&\textbf{19/50/100} (16\%)  \\
&Church & 0.9/9.1/87.9 & \textbf{1.9/14.6/99.1} (40\%) \\ \hline
\end{tabular}
$$
\label{tab:df_level}
\end{table}

We compare our method quantitatively with three mainstream single-frame non-UA APR methods and UA APR methods. We utilized the resulting similarity score to measure prediction uncertainty in this paper. Our \textit{uncertainty estimation module} applies to almost all APRs, and we present results over three different models. We gradually increase the threshold $\gamma$, filter predictions whose similarity scores are below $\gamma$, and analyze the average error of the remaining sample, as shown in Figure~\ref{fig:correlation}. The subfigures Figure~\ref{fig:correlation}.g-i indicate a decrease in the ratio of retained predictions as the threshold $\gamma$ increases, coupled with the decrease in normalized mean translation and rotation errors of the retained predictions as $\gamma$ increases in Figures~\ref{fig:correlation}.a-f.  This clear correlation demonstrates that samples with higher similarity scores tend to yield more accurate estimations.

\subsubsection{7Scenes}
For all three APR models, when the similarity threshold $\gamma$ is  higher than 0.95, the translation and rotation mean errors are reduced by 25\% to 80\% across the scenes.  

\subsubsection{Cambridge} For all three APR models, when $\gamma$ is higher than 0.95, the translation and rotation mean errors are reduced by 15\% to 80\% on different scenes. 

Non-UA APRs (MS-T and DFNet) exhibit higher accuracy individually compared to UA APRs (Bayesian PN, AD-PN, BMDN, and VaPoR) as shown in Table~\ref{tab:acc_7s_rank} and Table~\ref{tab:acc_cam_rank}. Our new APR-agnostic \textit{uncertainty estimation module} enables uncertainty estimation for such models, allowing us to leverage their greater accuracy in applications that require uncertainty measures.

\subsection{UA pose refinement evaluation}

Adding uncertainty awareness into APRs through the \textit{uncertainty estimation module} enables filtering unreliable poses and prioritizing the optimization of less accurate predictions.  

\subsubsection{7Scenes}
Table~\ref{tab:df_level} illustrates the performance improvement provided by solely filtering unreliable poses identified by the pose estimation module. DFNet$^{hr}$+Filter achieves higher percentages in all accuracy levels by keeping the $14\sim75\%$ predictions with reasonably high similarity scores pass $\gamma$. Furthermore, we completely filter out the pose predictions in the low accuracy level ($5m, 10^\circ$) in six of seven scenes, and reduce it to 0.1\% in the Office scene.
Regarding pose optimization, Figure~\ref{fig:bar_7s} shows that high similarity (hs) predictions, with similarity scores above $\gamma$, largely converge within 10 steps of optimization using the NeFeS method. On the other hand, low similarity (ls) predictions still require 10 to 50 steps to converge.

\begin{table*}[h]
\caption{Comparisons on 7-Scenes dataset.  The median translation and rotation errors (m/$^\circ$) of different methods and the average refine steps for each query image. The best results are in bold (lower is better).} 
\centering
\setlength{\tabcolsep}{4pt}
\begin{threeparttable}
\begin{tabular}{l|c|ccccccc|lc}
\hline    & Methods& Chess  & Fire  & Heads & Office &Pumpkin  & Redkitchen& Stairs  & Avg. [$m/^\circ$] &  Avg. Steps\\
\hline 
UA APR &Bayesian PN~\cite{kendall2016modelling}& 0.37/7.24 &0.43/13.7 & 0.31/12.0 & 0.48/8.04& 0.61/7.08& 0.58/7.54 & 0.48/13.1 & 0.47/9.81 & -\\
&CoordiNet~\cite{moreau2022coordinet}&0.14/6.7&0.27/11.6&0.13/13.6&0.21/8.6&0.25/7.2& 0.26/7.5&0.28/12.9&0.22/9.70& -\\
&BMDN\tnote{1}~\cite{bui20206d} & 0.10/6.47 &0.26/14.8 & 0.13/13.4 & 0.19/9.73&  0.20/9.40& 0.19/10.9&0.34/14.1 & 0.20/11.26& -\\
&VaPoR~\cite{10160466} & 0.17/6.9 & 0.30/14.1&0.17/14.5 &0.24/9.3&0.30/8.3 & 0.26/10.2&0.47/15.5 &0.27/11.26& -\\\hline
non-UA APR & PN~\cite{kendall2015posenet} & 0.10/4.02&0.28/9.44& 0.18/13.9& 0.17/5.99& 0.22/5.18& 0.23/5.91& 0.34/11.6& 0.22/7.16& -\\
& MS-T~\cite{shavit2021learning} & 0.11/6.45 &0.23/10.9 &0.13/13.1& 0.18/8.18& 0.16/6.87 & 0.17/8.44 & 0.30/10.4 & 0.18/9.19& -\\
 & DFNet~\cite{chen2022dfnet}  & 0.03/1.12&  0.06/2.30&  0.04/2.29 & 0.06/1.54&  0.07/1.92&  0.07/1.74 & 0.12/2.63 & 0.06/1.93& -\\\hline
APR Refine  & DFNet + NeFeS10~\cite{chen2023refinement} & \textbf{0.02/0.55} & 0.03/1.20 &0.03/1.88& 0.04/1.08 & 0.04/0.98 & 0.03/0.97 & 0.10/2.47 & 0.04/1.30& 10\\
&DFNet + NeFeS50~\cite{chen2023refinement} &\textbf{0.02}/0.57& \textbf{0.02/0.74} &\textbf{0.02/1.28}& \textbf{0.02/0.56} &\textbf{0.02/0.55} &\textbf{0.02/0.57} &\textbf{0.05/1.28}& \textbf{0.02/0.79}& 50\\\hline
Ours & DFNet$^{hr}$ + NeFeS$^{10}_{50}$ & \textbf{0.02/0.55}  & \textbf{0.02}/0.75 & \textbf{0.02}/1.45 & \textbf{0.02}/0.64  & \textbf{0.02}/0.62 & \textbf{0.02}/0.67 & \textbf{0.05}/1.30 & \textbf{0.02}/0.85&36.3 \\
\hline
\end{tabular}
\begin{tablenotes}
        \footnotesize
        \item[1] Results of BMDN taken from \cite{10160466}.
\end{tablenotes}
\end{threeparttable}
\label{tab:acc_7s_rank}
\end{table*}

\begin{table}[h]
\centering
\caption{Comparisons on Cambridge dataset.  The median translation and rotation errors (m/$^\circ$) of different methods and the average refine steps for each query image. Best results are in bold (lower is better).}
\setlength{\tabcolsep}{1pt} %
\begin{threeparttable}
\resizebox{\columnwidth}{!}{
\begin{tabular}{l|c|cccc|cc}
\hline
 &Methods & Kings  & Hospital  & Shop  & Church  &Avg. & Avg. Steps  \\\hline
UA APR& Bayesian PN & 1.74/ 4.06 &2.57/ 5.14&1.25/ 7.54  &2.11/ 8.38& 1.92/6.28 &-\\
&AD-PN & 1.3/1.67 & 2.28/4.80& 1.22/6.17&-/-  & -/- &-\\
&BMDN\tnote{1} & 1.51/2.14 & 2.25/3.93 & 3.52/5.41 & 2.16/5.99 & 2.36/4.37&-\\
&VaPoR & 1.65/2.88 & 2.06/4.33 & 1.02/6.03 & 1.80/5.90 & 1.63/4.79 &-\\
\hline
non-UA APR&PN & 0.93/2.73 & 2.24/7.88 & 1.47/6.62 & 2.37/5.94 & 1.75/5.79 &-\\
&MS-T  & 0.85/1.45 &1.75/2.43& 0.88/3.20 & 1.66/4.12 & 1.29/2.80 &-\\
&DFNet & 0.73/2.37 & 2.00/2.98 & 0.67/2.21 & 1.37/4.02 & 1.19/2.90 &-\\\hline
APR Refine&DFNet + NeFeS30& 0.37/0.64& 0.98/1.61 & 0.17/0.60 & 0.42/1.38 & 0.49/1.06 &30\\
&DFNet + NeFeS50& 0.37/\textbf{0.54}& \textbf{0.52/0.88} &0.15/0.53& \textbf{0.37/1.14}& \textbf{0.35/0.77} &50\\\hline
Ours &DFNet$^{hr}$ + NeFeS$^{30}_{50}$ & \textbf{0.36}/0.58&0.53/0.89 & \textbf{0.13/0.51} & 0.38/1.16 & \textbf{0.35}/0.78&42.4\\
\hline 
\end{tabular}
}
\begin{tablenotes}
\footnotesize
\item[1] Results of BMDN taken from \cite{10160466}.
\end{tablenotes}
\end{threeparttable}

\label{tab:acc_cam_rank}
\end{table}

\subsubsection{Cambridge}
\begin{figure}[h]
 \centering
\subfloat[Hospital (DFNet$^{hr}$)]{
 \label{fig:subfig:f} 
 \includegraphics[width=.3\linewidth]{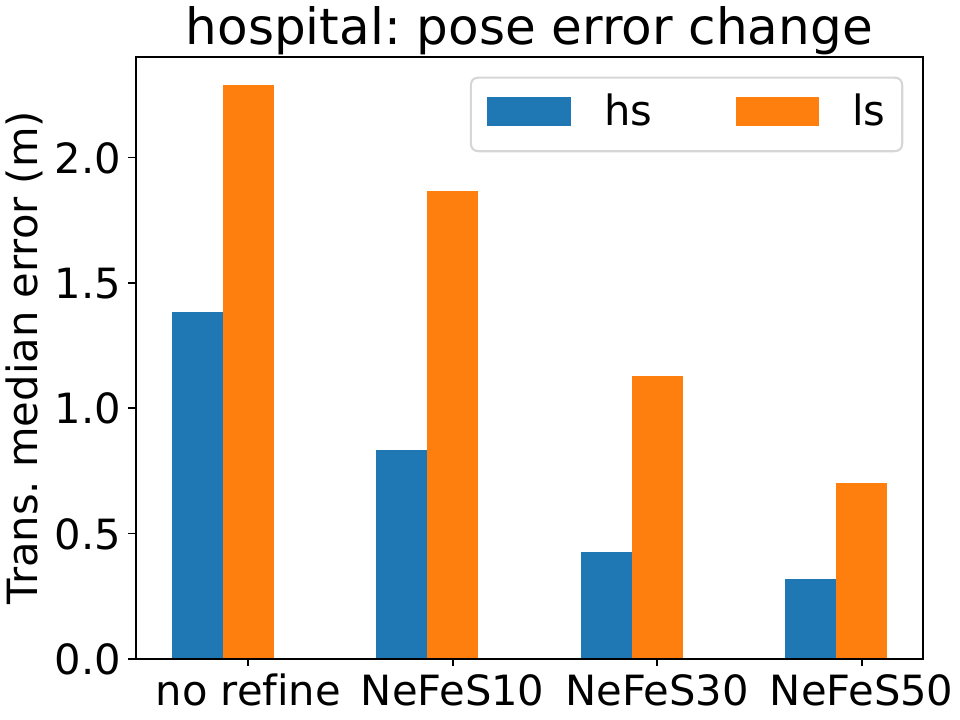}}
  \subfloat[Shop (DFNet$^{hr}$)]{
 \label{fig:subfig:f} 
 \includegraphics[width=.3\linewidth]{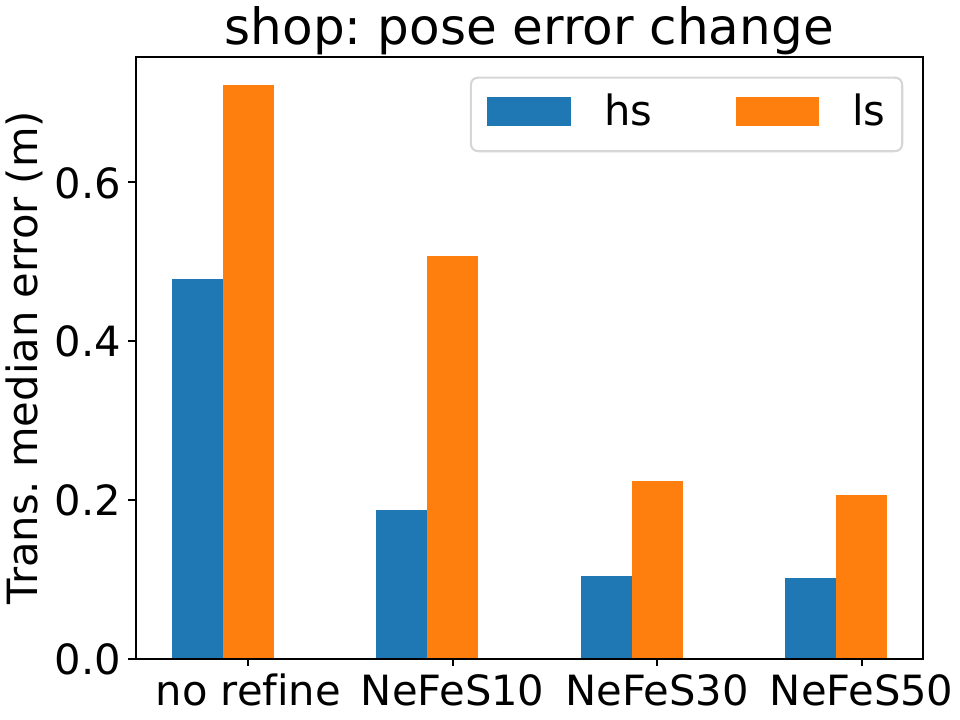}}
     \subfloat[Church (DFNet$^{hr}$)]{
 \label{fig:subfig:f} 
 \includegraphics[width=.3\linewidth]{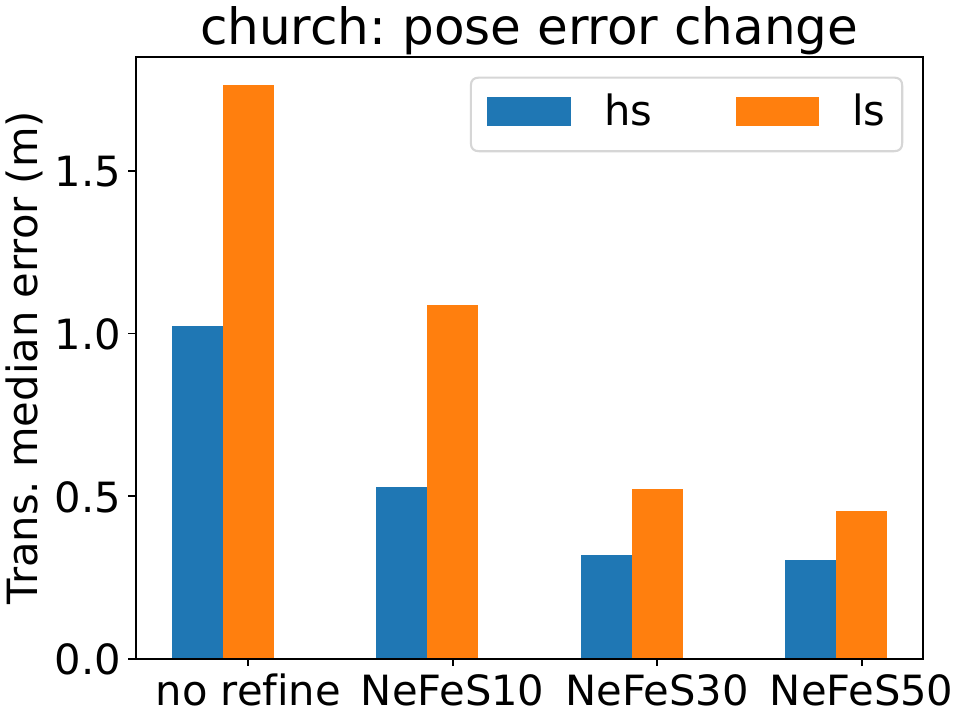}}
 \newline
 \subfloat[Hospital (DFNet$^{hr}$)]{
 \label{fig:subfig:a} 
 \includegraphics[width=.3\linewidth]{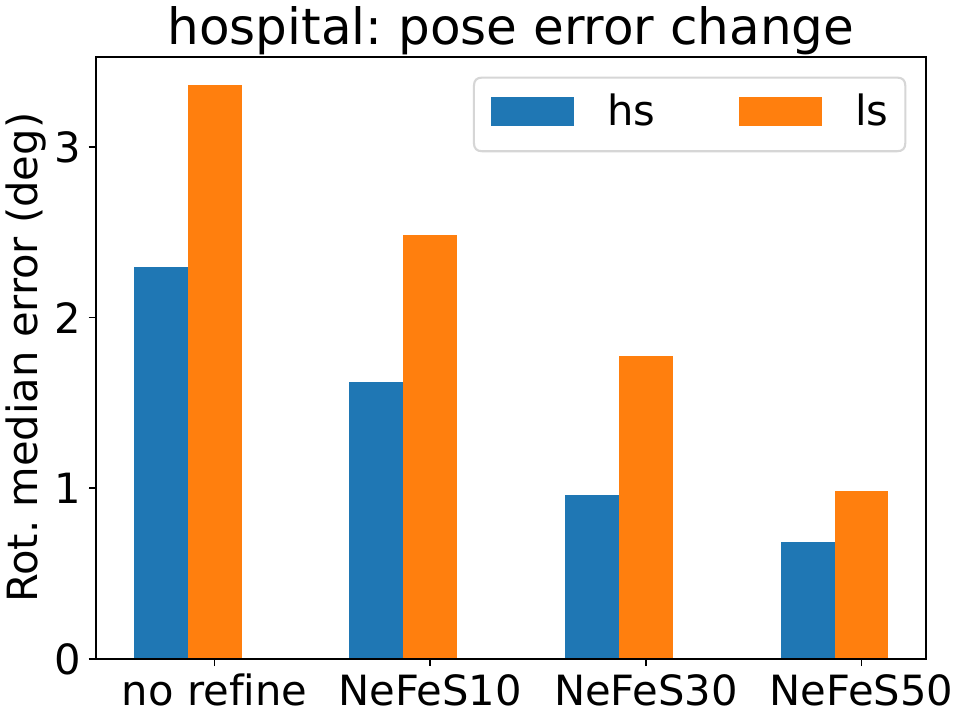}}
 \subfloat[Shop (DFNet$^{hr}$)]{
 \label{fig:subfig:a} 
 \includegraphics[width=.3\linewidth]{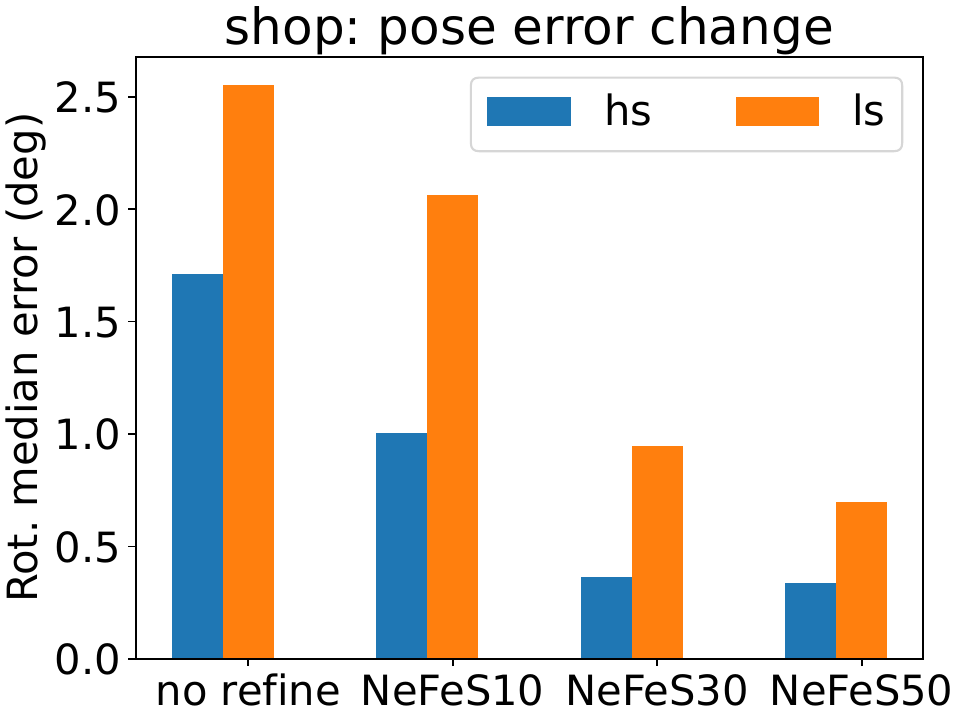}}
 \subfloat[Church (DFNet$^{hr}$)]{
 \label{fig:subfig:a} 
 \includegraphics[width=.3\linewidth]{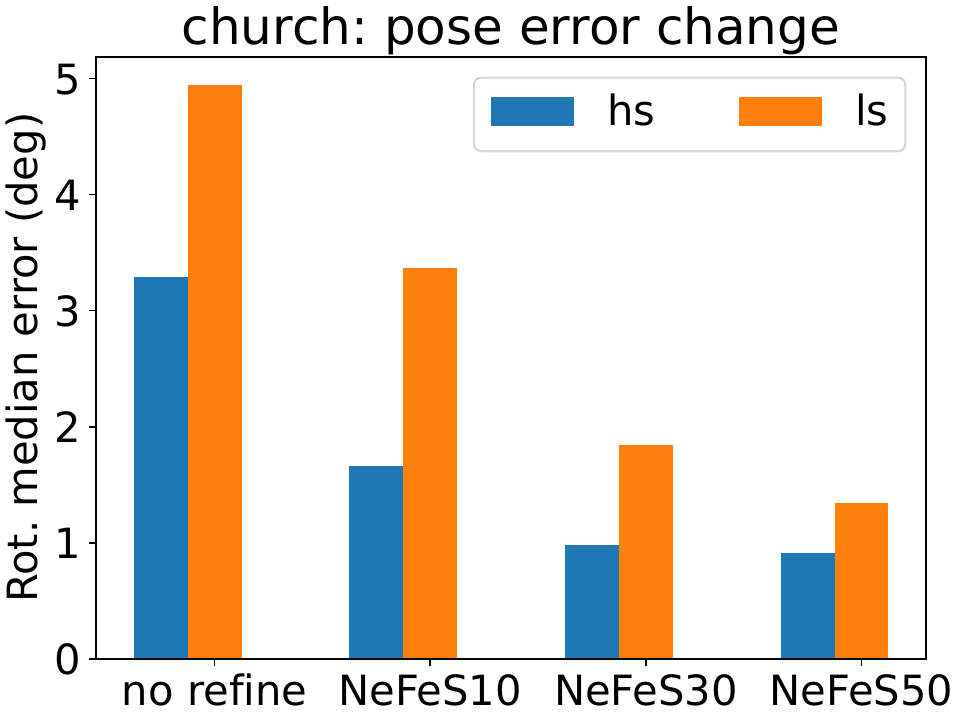}}
 \caption{Plots of translation and rotation errors against the
number of iteration for images pass the similarity threshold (hs) and images with low similarity scores (ls) do not pass the similarity threshold on some scenes of Cambridge. The hs predictions ratio of each scene is provided in Table~\ref{tab:df_level}. NeFeS$m$ denotes runing the refinement process for $m$ iterations.}
 \label{fig:bar_cam} 
\end{figure}

Similar to 7Scenes, Table~\ref{tab:df_level} shows that filtering unreliable poses achieves higher percentages in all accuracy levels by keeping the $13\sim55\%$ predictions with reasonably high similarity scores pass $\gamma$. Besides, we filter out the pose predictions in low accuracy level ($5m, 10^\circ$) in three of four scenes and reduce it to $0.9\%$ in the Church. As shown in Figure~\ref{fig:bar_cam}, high similarity (hs) predictions, with similarity scores above $\gamma$, largely converge within 30 steps of optimization using the NeFeS method, but low similarity (ls) predictions still require 30 to 50 steps to converge.

Table~\ref{tab:acc_7s_rank} and Table~\ref{tab:acc_cam_rank} show the accuracy of our UA pose refinement method compared to the SOTA non-UA APRs and refinement pipeline. The results demonstrate that our method achieves comparable accuracy to SOTA methods while reducing the optimization overhead by 
27.4\% and 15.2\% on the indoor and outdoor datasets, respectively.

\subsection{Analysis}
\begin{figure}[!t]
 \centering
  \subfloat[Chess ($\text{DFNet}^{hr}$)]{
 \label{fig:subfig:f} 
 \includegraphics[height=0.8in,width=0.45\linewidth]{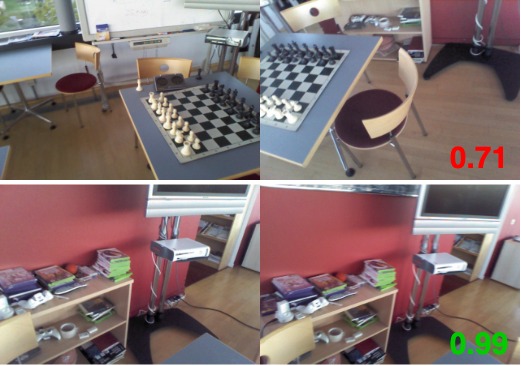}}
 \subfloat[Kings ($\text{DFNet}^{hr}$)]{
 \label{fig:subfig:a} 
 \includegraphics[height=0.8in,width=0.45\linewidth]{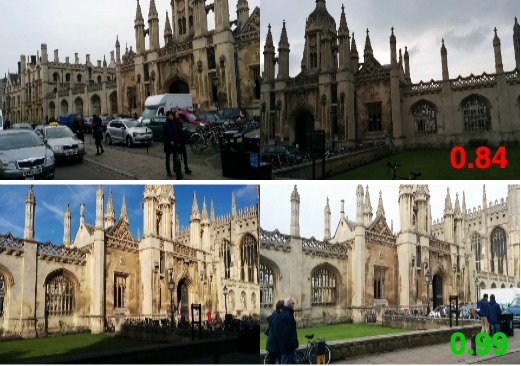}}
 \newline
\subfloat[Pumpkin ($\text{DFNet}^{hr}$)]{
 \label{fig:subfig:f} 
 \includegraphics[height=0.8in,width=0.45\linewidth]{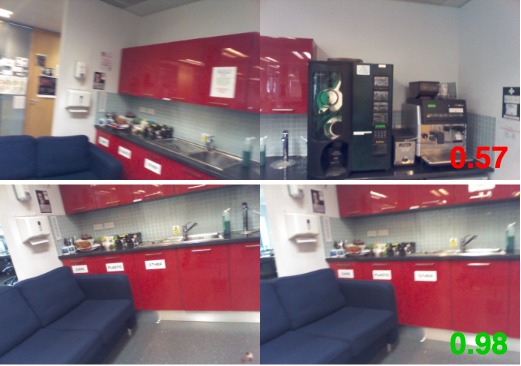}}
\subfloat[Church ($\text{DFNet}^{hr}$)]{
 \label{fig:subfig:f} 
 \includegraphics[height=0.8in,width=0.45\linewidth]{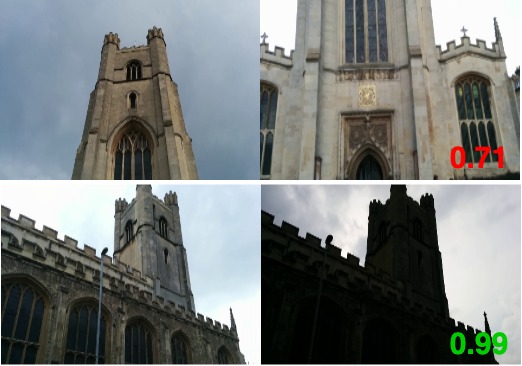}}
 \caption{Examples of retrieved image pairs in Cambridge and 7Scenes. For each pair, left side is the query image in test set, right side is the training set image retrieved by pose-based algorithm with the maximum similarity. The red numbers in the upper row of each subfigure represent the two images with low similarity and the green numbers in the lower row represent the two images with high similarity.}
 \label{fig:good_bad} 
\end{figure}
Figure~\ref{fig:track} shows that all three APR models have higher accuracy on query images closer to the training set's viewpoints.  Without knowing the ground truth of the query image, our uncertainty estimation module utilizes image features to measure the similarity between the query image and the training set, as we expect the similarity of image features to equate the proximity of viewpoints. We visualize the query image and the corresponding image in the training set retrieved by our pose-based
algorithm with the maximum similarity. As shown in Figure~\ref{fig:good_bad}, some query images are very similar to the training set images, while others are less straightforward to match. The clear correlation in Figure~\ref{fig:correlation} and improvement in all accuracy levels after filtering in Table~\ref{tab:df_level} confirm our uncertainty estimation module is effective and fits the design outcomes in Section~\ref{subsec:outcomes}. Besides, Figure~\ref{fig:good_bad}.b and Figure~\ref{fig:good_bad}.d show that the high-level and low-dimensional feature embeddings extracted from PoseNet-based feature extractor $E$ are robust to different lighting, weather, moving objects, and pedestrians where point-based SIFT registration fails~\cite{kendall2015posenet}.

\subsection{System Efficiency}
We evaluate the processing time of the proposed framework on a PC equipped with an NVIDIA GeForce GTX 3090 GPU. Based on the implementation in Section~\ref{sec:method} and Section~\ref{sec:exp}, we repeat each measurement 1000 times. The feature extraction of $E$ takes $3.4ms$. Pose-based feature retrieval with $\mathcal{O} (n)$ complexity takes $1.98ms$, while similarity calculation takes $1.3ms$ in the Heads dataset. Overall, the \textit{uncertainty estimation module} only adds about an extra $6.7ms$ to each inference time of APRs.  That means predictions can get the uncertainty in real-time. The feature extractor $E$ with EfficientNetb0 as backbone only takes 20 MB to store the weights, and each $1\times1024$ feature embedding of an image only takes 4.2 KB. That means our \textit{uncertainty estimation module} is also very storage efficient. Through its high modularity, our framework provides a new paradigm for real-life camera relocalisation with APRs. It can integrate most existing and future APRs and test-time refinement methods with minimal changes.

%% file: Sections/5Conclusion.tex
\section{CONCLUSIONS}
We introduce an APR-agnostic framework, \sysname, which includes an uncertainty estimation module that uses a novel pose retrieval algorithm to calculate the cosine similarity between image feature embeddings in the training set and the query image. This uncertainty estimation is used to optimize an iterative pose refinement algorithm. We evaluate the performance of our method on indoor and outdoor datasets using three different APRs. Our results demonstrate a clear correlation between pose error and uncertainty, validating the effectiveness of our approach. Moreover, \sysname significantly reduces the computational overhead of the refinement pipeline while maintaining accuracy.